\documentclass[sigconf]{aamas}

\usepackage{balance} 
\usepackage[ruled,noend]{algorithm2e}

\usepackage{fancyhdr}
\definecolor{brown}{rgb}{.69,.37,.22}

\newcommand{\frombody}[3]{
	\noindent
	\textcolor{#1}{
		{$\bf [\!\![\!\![$}\underline{\scshape{#2}}
		{\scshape says:} 
		\textsl{#3}{$\bf ]\!\!]\!\!]$}}
	{}}
\renewcommand{\frombody}[3]{}

\newcommand{\secref}[1]{\S\ref{#1}}
\usepackage{amsmath}
\newcommand{\exampledone}{\hfill \ensuremath{\blacksquare}}

\usepackage{xspace}

\newcommand{\etal}{\emph{et~al.}\xspace}

\newcommand{\zeroone}{\ensuremath{\{0,1\}}\xspace}
\newcommand{\zot}{\ensuremath{\{0,1,2\}}\xspace}

\newcommand{\mdpmaster}{\ensuremath{M}\xspace}
\newcommand{\mdpstates}[1]{%
   \ifthenelse{ \equal{#1}{} }
      {\ensuremath{\mathcal{S}}\xspace}
      {\ensuremath{\mathcal{S}_{#1}}\xspace}}
\newcommand{\mdpstatesi}{\mdpstates{i}}
\newcommand{\mdpstate}{\ensuremath{s}\xspace}
\newcommand{\mdpactions}{\ensuremath{\mathcal{A}}\xspace}
\newcommand{\mdpactionsi}{\ensuremath{\mdpactions_i}\xspace}
\newcommand{\mdpaction}{\ensuremath{a}\xspace}
\newcommand{\mdptransition}{\ensuremath{p}\xspace}
\newcommand{\mdptransitioni}{\ensuremath{\mdptransition_i}\xspace}
\newcommand{\mdpreward}[1]{
    \ifthenelse{ \equal{#1}{} }
    {\ensuremath{r}\xspace}
    {\ensuremath{r_{#1}}\xspace}}
\newcommand{\mdprewardi}{\mdpreward{i}}

\newcommand{\tasks}{\ensuremath{\mathcal{T}}\xspace}
\newcommand{\task}[1]{%
   \ifthenelse{ \equal{#1}{} }
      {\ensuremath{m}\xspace}
      {\ensuremath{m_{#1}}\xspace}}
\newcommand{\taski}{\task{i}}

\newcommand{\transitiondomain}{\ensuremath{\mathcal{D}^\tasks}\xspace}
\newcommand{\curriculum}{\ensuremath{C}\xspace}
\newcommand{\currvertices}{\ensuremath{\mathcal{V}}\xspace}
\newcommand{\curredges}{\ensuremath{\mathcal{E}}\xspace}
\newcommand{\taskdescriptor}{\ensuremath{\theta}\xspace}

\newcommand{\Proficiency}{\ensuremath{\Psi}\xspace}
\newcommand{\proficiency}{\ensuremath{\psi}\xspace}
\newcommand{\parentfunctions}{\ensuremath{\Phi}\xspace}

\newcommand{\option}[1]{%
   \ifthenelse{ \equal{#1}{} }
      {\ensuremath{o}\xspace}
      {\ensuremath{o_{#1}}\xspace}}

\newcommand{\oterm}{\ensuremath{\textsc{TERM}}}

\newcommand{\target}[1]{%
   \ifthenelse{ \equal{#1}{} }
      {\ensuremath{\kappa}\xspace}
      {\ensuremath{\kappa_{#1}}\xspace}}
\newcommand{\Target}{\ensuremath{K}\xspace}

\newcommand{\discount}{\ensuremath{\gamma}}

\usepackage{enumitem}

\makeatletter
\gdef\@copyrightpermission{
  \begin{minipage}{0.2\columnwidth}
   \href{https://creativecommons.org/licenses/by/4.0/}{\includegraphics[width=0.90\textwidth]{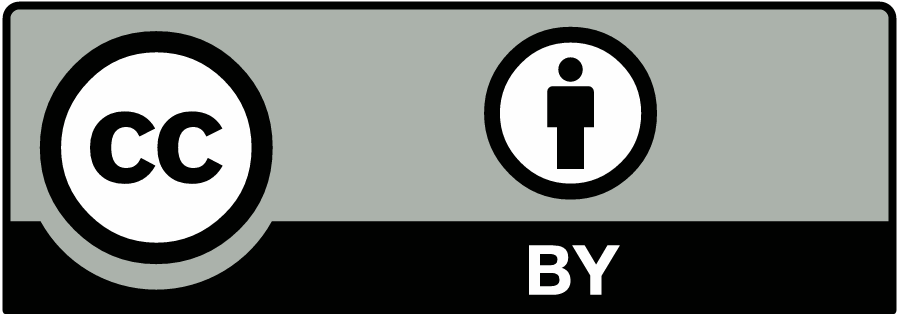}}
  \end{minipage}\hfill
  \begin{minipage}{0.8\columnwidth}
   \href{https://creativecommons.org/licenses/by/4.0/}{This work is licensed under a Creative Commons Attribution International 4.0 License.}
  \end{minipage}
  \vspace{5pt}
}
\makeatother

\setcopyright{ifaamas}
\acmConference[AAMAS '25]{Proc.\@ of the 24th International Conference
on Autonomous Agents and Multiagent Systems (AAMAS 2025)}{May 19 -- 23, 2025}
{Detroit, Michigan, USA}{Y.~Vorobeychik, S.~Das, A.~Nowé  (eds.)}
\copyrightyear{2025}
\acmYear{2025}
\acmDOI{}
\acmPrice{}
\acmISBN{}

\acmSubmissionID{950}

\title[AAMAS-2025 Formatting Instructions]{
Automating Curriculum Learning for Reinforcement Learning using a Skill-Based Bayesian Network
}

\author{Vincent Hsiao}
\affiliation{
  \institution{NRC Postdoctoral Fellow, \\Naval Research Laboratory}
  \city{Washington DC}
  \country{United States}}
\email{vincent.hsiao.ctr@us.navy.mil}

\author{Mark Roberts}
\affiliation{
  \institution{Naval Research Laboratory}
  \city{Washington DC}
  \country{United States}}
\email{mark.c.roberts20.civ@us.navy.mil}

\author{Laura M. Hiatt}
\affiliation{
  \institution{Naval Research Laboratory}
  \city{Washington DC}
  \country{United States}}
\email{laura.m.hiatt.civ@us.navy.mil}

\author{George Konidaris}
\affiliation{
    \institution{Brown University}
    \city{Providence RI}
    \country{United States}}
\email{gdk@brown.edu}

\author{Dana Nau}
\affiliation{
    \institution{University of Maryland}
    \city{College Park, MD}
    \country{United States}}
\email{nau@umd.edu}

\begin{abstract}
A major challenge for reinforcement learning is automatically generating curricula to reduce training time or improve performance in some target task.
We introduce SEBNs (Skill-Environment Bayesian Networks)
which model a probabilistic relationship between a set of skills, a set of goals that relate to the reward structure, and a set of environment features to predict policy performance on (possibly unseen) tasks.  
We develop an algorithm that uses the inferred estimates of agent success from SEBN to weigh the possible next tasks by expected improvement.
We evaluate the benefit of the resulting curriculum on three environments: a discrete gridworld, continuous control, and simulated robotics.
The results show that curricula constructed using SEBN frequently outperform other baselines.
\end{abstract}


\newcommand{\BibTeX}{\rm B\kern-.05em{\sc i\kern-.025em b}\kern-.08em\TeX}

\begin{document}


\pagestyle{fancy}
\fancyhead{}


\maketitle 


\fancypagestyle{firststyle}
{
\cfoot{ \begin{small} \textbf{DISTRIBUTION STATEMENT A.} Approved for public release; distribution is unlimited. \end{small}}
}
\thispagestyle{firststyle}

\section{Introduction}
Adapting skills to new or unseen tasks is a major challenge in Reinforcement Learning (RL). Curriculum Learning \cite{bengio2009curriculum, narvekarEtAl20.jmlr.clForRL}, an approach for training agents using a sequence of increasingly difficult environments, often promotes the effective development of policies with more robust capabilities. However, customizing a curriculum to a particular student often requires substantial human insight and oversight. This is especially challenging for robotics, where the environment or tasks that need to be performed can change frequently. An ideal solution to this problem would be an automated curriculum that enables the robot to discern for itself when it needs to adapt, how long should train, and in what environments. 

Past work on automated curriculum generation such as \cite{stout2010competence, kumar2024practice} has primarily focused on choosing what skills to train while holding the environment itself static. More recent approaches that build a curriculum over different environments such as \cite{parker2022evolving} do not consider agent skill competencies. Furthermore, these environment-based approaches require explicit evaluation on an environment before being able to calculate an estimate of agent success or regret to add those environments to the curriculum. 

We address the aforementioned issues by introducing Skill Environment Bayesian Networks (SEBNs) as a potential method for estimating agent competency level and selecting the most appropriate environments for training. 
SEBNs model a probabilistic relationship between these goals, (latent) competencies, and environment features using data from past rollouts. Using this model, we can estimate agent success rates on new (possibly unseen) environments. We use these estimates to select the next set of training tasks within a curriculum in what we call an SEBN-guided automated curriculum. Importantly, SEBN does not require explicit evaluation on each possible environment to estimate agent success.

The contributions of the paper include:
    (1) Introducing and formalizing the SEBN for skills, task features, and reward structure;
    (2) Providing an algorithm for constructing curricula using SEBNs; 
    (3) Introducing  Megagrid, a gridworld environment that simplifies generating partially-specified environments for transfer learning; 
    (4) Assessing SEBN-based curricula on three distinct environments: a discrete gridworld (DoorKey), continuous control (BipedalWalker), and a difficult simulated robotics domain (robosuite); and
    (5) Demonstrating via experiments that SEBN curricula produce more robust policies that reach success more quickly than other curricula in the continuous control and robotics environments, and performed comparably in the gridworld environment.

\section{Background and Preliminaries}

Bayesian statistics rely on some sort of informed prior, provided or learned, to estimate the future values.  
In an SEBN, part of this prior is provided in the form of the network and in the strength of relationships, and part of the prior is learned through the collection of samples from the environment.
We next provide our motivation for this Bayesian approach (\secref{sec:ecd})
with some background on Bayesian Networks (\secref{sec:bns}).  
We then formalize the curriculum learning problem (\secref{sec:curriculum-learning}), how we use task features to construct tasks (\secref{sec:taskdescriptors}), and an extension to the options framework (\secref{sec:options}).

\subsection{Evidence Centered Design}
\label{sec:ecd}

Our motivation for using a Bayesian Network to estimate learning proficiency comes from the method of Evidence Centered Design (ECD) \cite{mislevy2003brief}, a technique used in human educational assessments. 
In Evidence-Centered Design, statistical models, such as Bayesian networks, are used to measure the proficiency levels of a given student. These proficiency measurements are then used to inform task and assessment creation. For example, ECD could be used to help model and analyze the performance of a tennis player. The Bayesian network in this domain can include nodes that represent latent competencies (e.g., mobility, footwork, dynamic vision, etc.) and nodes that represent observable metrics (e.g., number of successful serves, return rate, game score). The performance of a tennis player on the observable metrics is used to infer their latent capabilities. New training goals can then be set using these estimated capabilities. This technique is effective in human educational contexts, and we hypothesize that a similar approach could be applied to assist in designing a curriculum to improve learning in robotic agents. 

\subsection{Bayesian Networks (BNs)}
\label{sec:bns}

Bayesian Networks (BNs) \citep{pearl} are a type of graphical model that provide an efficient way to represent and reason about probabilistic relationships among a set of random variables. A Bayesian Network $(X,D,\parentfunctions)$ is defined by a set of variables $X$, their corresponding domains $D$, and a set of parent functions \parentfunctions that specify the conditional probability distributions of each variable given its parents. 
When $D$ is discrete, these parent functions are typically specified in a tabular format known as Conditional Probability Tables (CPTs). 

It is common to use BNs to model relationships between latent and observable variables. 
Once constructed, the network can be used to infer latent values from observed values. 
New data can be entered in the form of evidence values for observed variables in a BN. 
The probabilities over other variables in the network are calculated by conditioning on this observed evidence, i.e., as a conditional probability: $P(X_1|X_2,...,X_N)$. 
We will employ a standard bucket elimination algorithm (aka variable elimination) \cite{darwiche2009modeling, dechter2013reasoning} to perform inference.
In this paper, the observable variables of the BN relate to the environment of an agent and a target performance it is attempting to achieve; both are modeled as a task in an MDP problem and defined in \secref{sec:curriculum-learning}.  The unobservable variables will relate to a set of latent competencies that we define in \secref{sec:latent-skills}. 

\subsection{Curriculum Learning}
\label{sec:curriculum-learning}

We adapt the notation of 
\citeauthor{narvekarEtAl20.jmlr.clForRL}~\cite{narvekarEtAl20.jmlr.clForRL} to describe a \emph{task} as the interaction of an agent with its environment to meet some objective.  
A task, formalized as an episodic Markov Decision Process (MDP), is a tuple 
$\mdpmaster = (\mdpstates, \mdpactions, \mdptransition, \mdpreward{})$, where 
    \mdpstates{} 
    is the set of states, 
    \mdpactions~is the set of actions, 
    $\mdptransition(\mdpstate'|\mdpstate, \mdpaction)$ gives the probability of being in state $\mdpstate'$ after taking action \mdpaction in state \mdpstate, and 
    $\mdpreward{}(\mdpstate, \mdpaction, \mdpstate')$ is the reward function after taking action \mdpaction in state \mdpstate and transitioning to state $\mdpstate'$.
A solution to \mdpmaster is a policy $\pi$ that maximizes the cumulative sum of rewards for an episode of length T, i.e., $\sum_{t=1}^{T} R_{t}$.

Let \tasks be a set of all tasks an agent could complete in \mdpmaster, where a task $\taski \in \tasks$ is a task-specific MDP $\taski = (\mdpstatesi, \mdpactionsi, \mdptransitioni, \mdprewardi)$.  For all tasks in \tasks, let \transitiondomain be the set of all possible transition samples from \tasks (see \citeauthor{narvekarEtAl20.jmlr.clForRL} \cite{narvekarEtAl20.jmlr.clForRL} for a complete definition).  
In their formalism, a Curriculum $\curriculum = (\currvertices, \curredges, \taskdescriptor, \tasks)$ 
is a directed acyclic graph, where \currvertices is the set of vertices, $\curredges \subseteq \{(x,y) | (x,y) \in \currvertices \times \currvertices \land x \neq y\}$ is a set of directed edges,  $\taskdescriptor: \currvertices \rightarrow \mathcal{P}(\transitiondomain)$ is a function that associates samples within each vertex, and $\mathcal{P}(\transitiondomain)$ is the powerset of \transitiondomain.

In this paper, we develop what Narvekar~\etal~\cite{narvekarEtAl20.jmlr.clForRL} call a task-level curriculum, where each vertex $v \in \currvertices$ is associated with samples from a single task in \tasks.  
That is, the mapping function for task \taski is 
$\taskdescriptor: \currvertices \rightarrow \{ \transitiondomain_i | \taski \in \tasks \}$.  
For convenience, we will refer to a task's available samples at vertex $v$ as $\taski^{\taskdescriptor}$.
In other words, a task descriptor $\taskdescriptor_i$ is used to construct task \taski, and a curriculum is a sequence of tasks \task{1}, \task{2}, ..., \task{target} up to some target task.

Before we describe how we construct this function using task features, we point out some deviations from the model just described.  
The curriculum being a DAG is a very strong assumption and is not true of the SEBN-guided curriculum.  
While the episodic MDP model of \citeauthor{narvekarEtAl20.jmlr.clForRL} provides a more comprehensible model of curriculum learning, the RL algorithms of this paper actually learn with a discount factor \discount, and one could argue that the Partially Observable MDP might be more appropriate.  Both changes would be extensions to the simplified MDP model presented here. 


\subsection{Task Descriptors (Env't+Target Features)}
\label{sec:taskdescriptors}
We will use \emph{task features} to define \taskdescriptor for a task $\taski^{\taskdescriptor}$.
This is a common approach to quantify potential transfer between two tasks (e.g., \cite{rostamiEtAl20.jair.usingTaskDescriptions,isele2016task,narvekar2016source,konidaris2012transfer}).
The notion is that two tasks that share similar features will exhibit better transfer.
We adopt a task descriptor similar to \citeauthor{rostamiEtAl20.jair.usingTaskDescriptions}~\citep{rostamiEtAl20.jair.usingTaskDescriptions} and \citeauthor{narvekar2016source}~\citep{narvekar2016source}.

Specifically, we parameterize \taskdescriptor with a vector that 
consists of set of environment-specific features $E$ and a set of one or more performance targets $\Target$.
Thus, $\taskdescriptor((\mathbb{Z}_0)^{|E|}(\mathbb{Z}_0)^{|K|})$ will indicate the specific task $\taski^{\taskdescriptor}$.
We will often omit the task descriptor for clarity and just reference \taski.  Note that the task descriptor is underspecified with respect to the environment, so one configuration of \taskdescriptor represents a class of different environments an agent can encounter.

\begin{figure}
\begin{scriptsize}
\begin{tabular}{ll}
    \includegraphics[trim={0 16cm 0 0}, clip, width=0.20\textwidth]{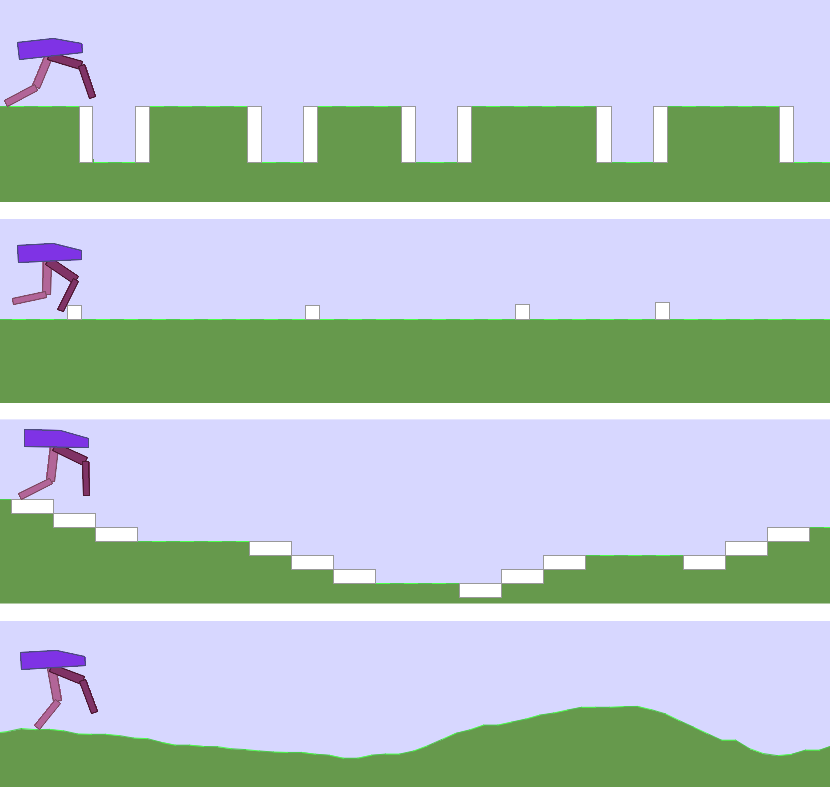} &     
    \includegraphics[trim={0 4.5cm 0 11cm}, clip, width=0.20\textwidth]{figures/bipedalwalkerenv.png}  \\
    (P3 S0 N0 W0 R0) & 
    (P0 S0 N3 W3 R0)\\
    \includegraphics[trim={0 10cm 0 6.3cm}, clip, width=0.20\textwidth]{figures/bipedalwalkerenv.png} & 
    \includegraphics[trim={0 0 0 16cm}, clip, width=0.20\textwidth]{figures/bipedalwalkerenv.png}\\
    (P0 S1 N0 W0 R0) & 
    (P0 S0 N0 W0 R4)\\
\end{tabular}
\end{scriptsize}
    \caption{Challenge environments for BipedalWalker  with corresponding descriptors (P:pit gap, S:stump height, W:stair width, N:stair steps, and R:ground roughness).  }

    \label{fig:bipedalwalkerenv}
\end{figure}

\begin{figure}
    \includegraphics[width=0.4\textwidth]{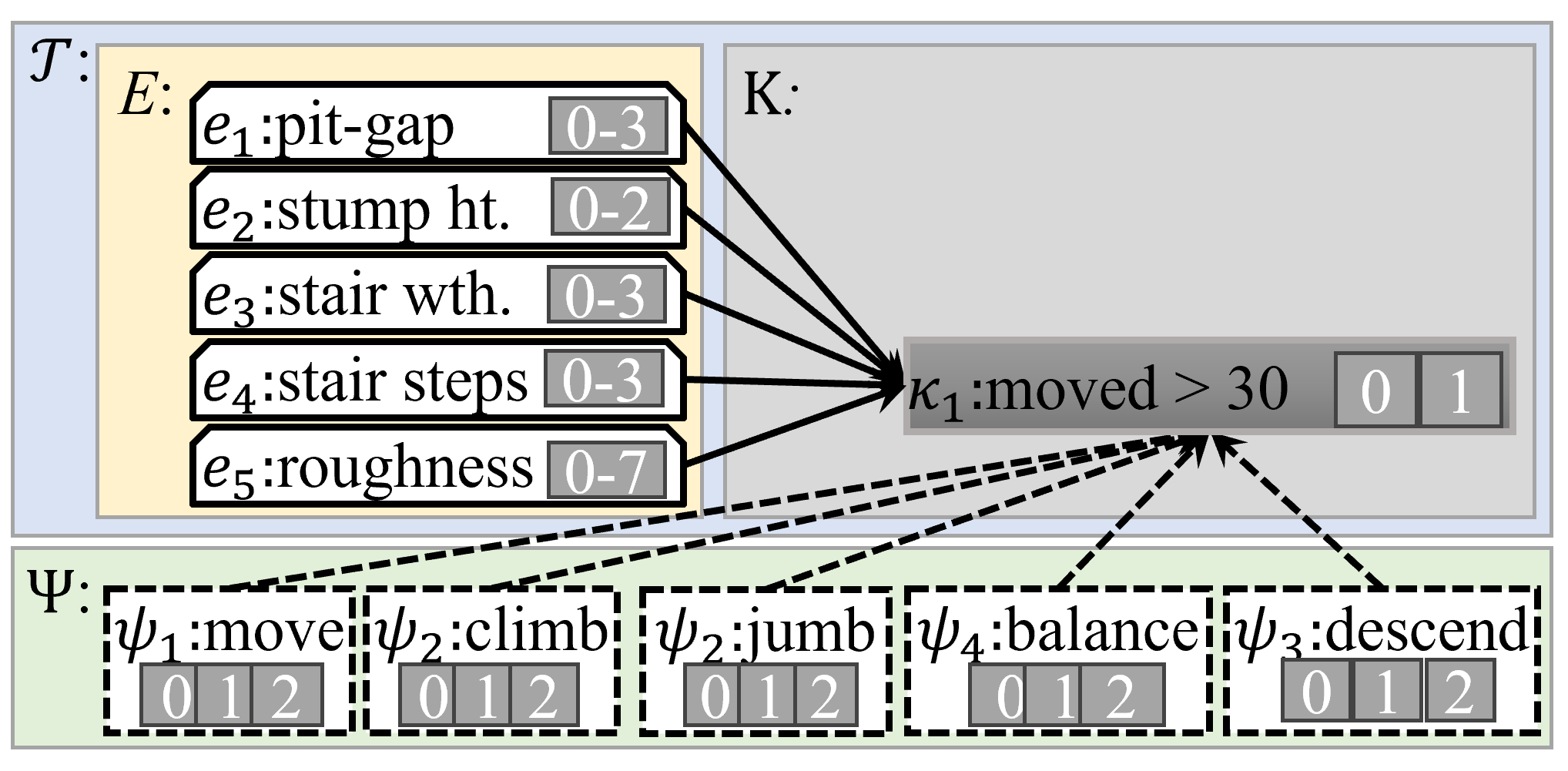}
    \caption{The SEBN for the Bipedal Walker environment.  }
    \label{fig:sebn-bipedal-walker}
\end{figure}

\paragraph{Example Task Descriptors using Bipedal Walker.} 
The Bipedal Walker (BPW) benchmark \cite{towers_gymnasium_2023} involves two-legged agent moving through terrain in a 2D environment.  Figure~\ref{fig:bipedalwalkerenv} shows some example terrains.
The top portion Figure~\ref{fig:sebn-bipedal-walker} shows $E$ and \Target for the bipedal walker. 
Here, $E$ consists of five features that control the difficulty of the environment, and there is a single target of moving to the right by at least 30 steps (The dashed latent competencies are defined in \secref{sec:latent-skills}). 
The task descriptor for this BPW is $\taskdescriptor($\mbox{\scriptsize{pit-gap, stump-height, stair-width, stair-steps, roughness, moved>30}}$)$. A task $\taski^{\taskdescriptor}$ for BPW involves a particular setting of the parameters for \taskdescriptor. \exampledone

\subsection{Extending Targets to Include Options}
\label{sec:options}

\noindent
One could imagine a richer set of targets in $K$, even ones that are hierarchically organized with a natural decomposition of subtasks.
The options framework \cite{suttonEtAl99.aij.betweenMDPsAndSemiMDPs}
is a common model for such situations.
Briefly, a subtask $j$ for task \taski is an option
$o_j = ( \mdpstates{j}, \pi_j, \oterm_j )$, 
where $\mdpstates{j} \subseteq \mdpstatesi$ are the starting states of the option, $\pi_j$ is used to take action while the option is enabled, and $\oterm_j: \mdpstatesi \rightarrow \zeroone$ is a function that indicates the option has terminated.

For a specific task \taski, a subtask  $\task{i}^j = (\mdpstates{j}, \mdpactionsi, \mdptransitioni, \mdpreward{j})$ indicates that an option has a specific context: it works over a set of states \mdpstates{j} that are a subset of the tasks states \mdpstatesi{}, it uses a specific reward independent of the task reward, and it has the same actions and transitions as the original task \taski.
Each option $j$ is enabled as part of the feature parameters for \taskdescriptor (i.e., $(\mathbb{Z}_0)^{|K|})$). 


\begin{figure}
\begin{scriptsize}
    
\begin{tabular}{p{1.8cm}p{1.8cm}p{1.8cm}p{1.8cm}}
\includegraphics[width=0.09\textwidth]{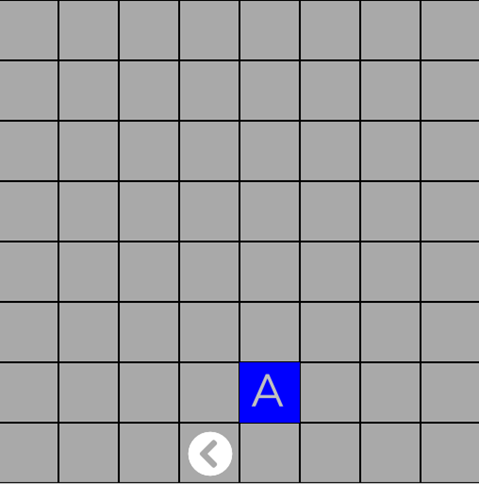} & 
\includegraphics[width=0.09\textwidth]{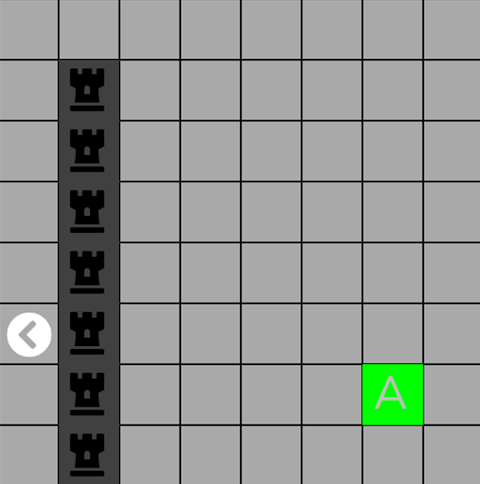} &
\includegraphics[width=0.09\textwidth]{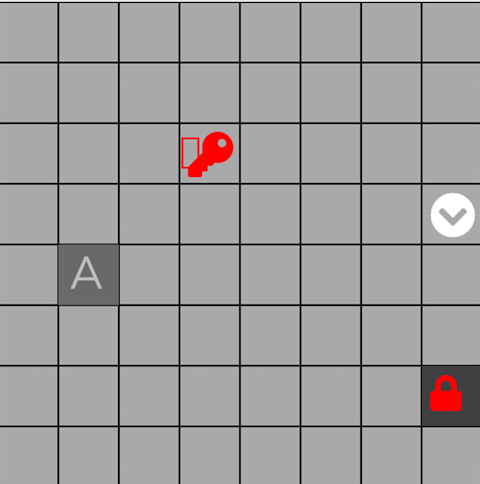} & 
\includegraphics[width=0.09\textwidth]{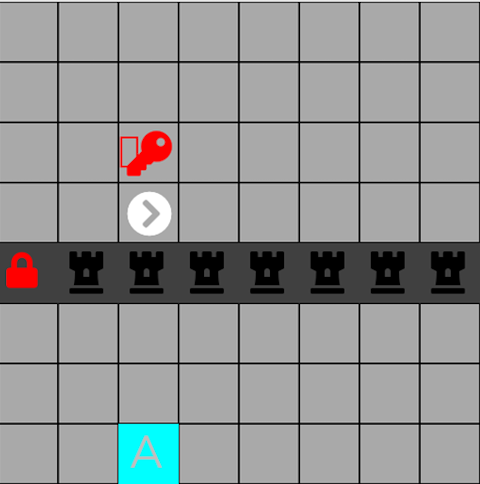} \\
(D0 W0 L0 K0 O0 A1) & (D1 W1 L0 K0 O0 A1) &
(D1 W0 L1 K0 O0 A1) & (D0 W1 L1 K1 O1 A1) \\

\end{tabular}
\end{scriptsize}
    \caption{Example environments for DoorKey with  corresponding environment features (D:distance, W:wall, L:locked door) and target features (K:key, O:opened, A:at).}
    \label{fig:doorkeyenv}
\end{figure}

\paragraph{Example Task Descriptors using DoorKey.}
Suppose we want an agent to learn to navigate in a gridworld environment to a goal while opening locked doors.
Fig.~\ref{fig:doorkeyenv} shows several possible environments for this agent and their corresponding environment feature vector. In the easiest environment (top left), the agent (the white arrow) starts very near the goal (``A'') in an empty grid. Adding additional obstacles such as a wall, shown as chess rooks, or a locked door, shown as a lock, with a key to unlock it, adjusts the environmental features accordingly. The first three components of the task descriptor  \taskdescriptor indicate whether the distance (D) of agent starts near (within 2 squares) of any point of interest (key or goal), the presence of a wall (W), and the presence of a locked door (L).

DoorKey also enables the use of options.  
The top right part of Figure~\ref{fig:doorkeysebn} shows a network of targets $K$ for this problem, corresponding to ordered subtasks.
This problem has three options (at(goal), opened(door), and has(key)), each with its own reward. 
A distinct policy is learned for each of these options.
The last three components of \taskdescriptor indicate which of these three subgoals are enabled for a task: getting a key (K), opening a door (O), or being at (A) a cell. \exampledone

\section{Bayesian Curriculum Learning}

The key idea in this section is to use a BN to estimate performance on $K$ over the environmental features from $E$ plus a set of estimated proficiency \Proficiency on latent competencies, which are hidden or unobserved.
Before we formally define the SEBN, we describe extend the DoorKey example to discuss this process.

\paragraph{Example of Latent Competencies}
Suppose we have collected past data of the agent's performance on different environments in $E$. For example, say we ran our agent on the empty-grid (D0 W0 L0), wall-only (D1 W1 L0), and door-only (D1 W0 L1) environments and recorded that the agent was successful on the empty-grid and door-only environments but not the wall-only environment. 
This failure might be due to the  agent not yet knowing how to navigate around walls. We can think of this ability as a latent "avoid wall" capability that an agent needs to have mastered to solve tasks that require it. Furthermore, using the notion that there is this latent capability, we can easily predict that the agent will fail on the wall-and-door (D0 W1 L1) environment without having any data of the agent's performance on that specific type of environment. \exampledone

\begin{figure}
    \includegraphics[width=0.35\textwidth]{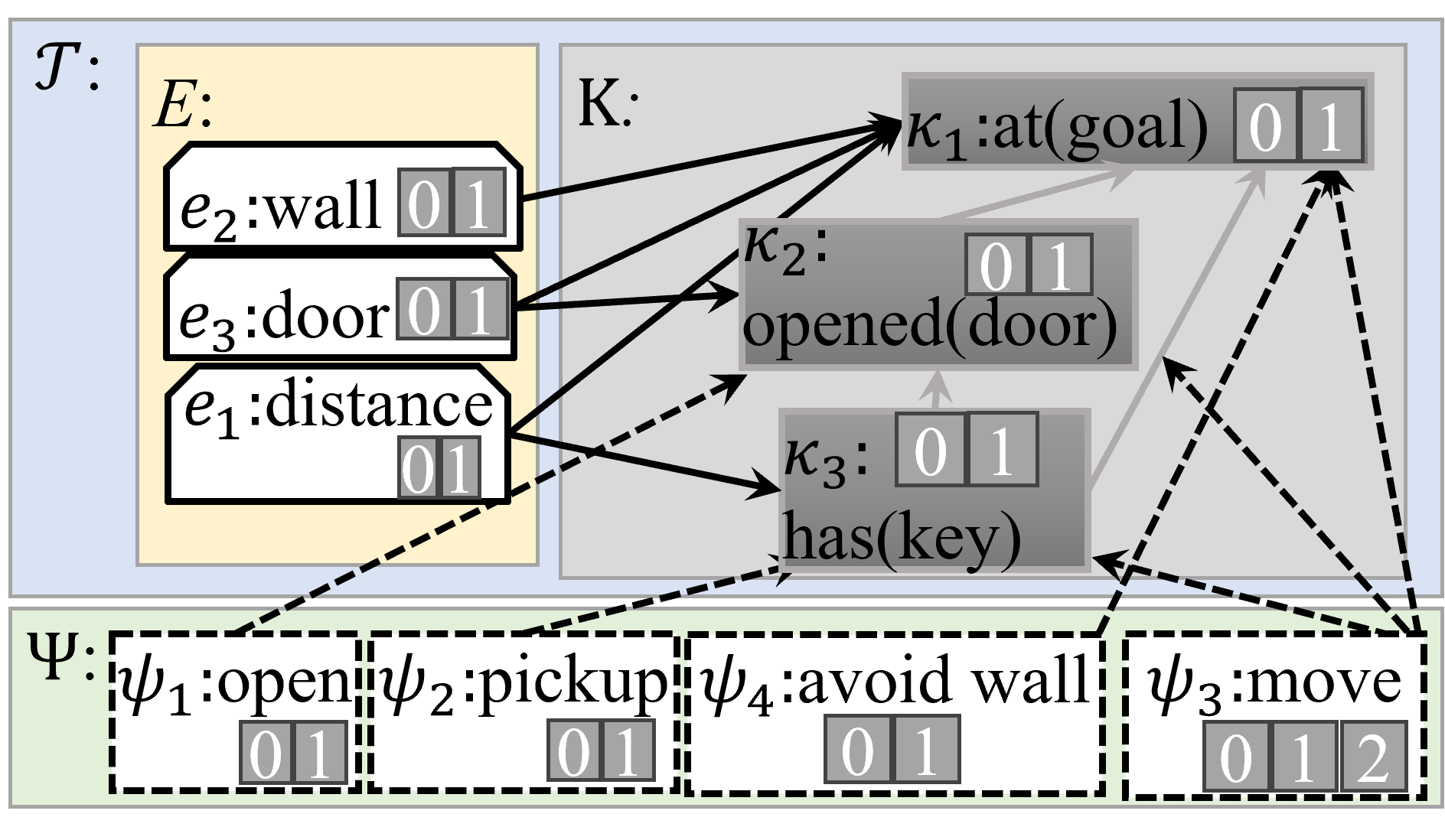}
    \caption{The SEBN for the DoorKey environment.  }
    \label{fig:doorkeysebn}
\end{figure}

The bottom row of Fig. \ref{fig:doorkeysebn} shows a set of latent competencies or capabilities \Proficiency. In this example, we chose four latent competencies: (move, pick up, avoid wall, open) that we expect the agent to need to master to successfully solve different tasks.  These latent variables are provided by an expert, similar to \citeauthor{abelEtAl15.icaps.goalBasedActionPriors} \cite{abelEtAl15.icaps.goalBasedActionPriors}.

We can use the SEBN to predict two important quantities. 
First, when faced with a new (possibly unseen) task $\taski \in \tasks$, we need to estimate the proficiency of each competency in \Proficiency.  This is important because competencies will advance at different rates and some tasks will require more proficiency than others for specific competencies.

Second, when faced with a new (possibly unseen) task, we need to be able to predict performance on $k_j \in K$ given the current estimates of competency level (from the first step). This is important because it can be used to select from a set of candidate tasks for training in the next iteration of a curriculum.



Returning to Fig.~\ref{fig:doorkeysebn}, suppose a new task is defined over E and \Target.  The proficiency estimate(s) can be calculated using the links to the bottom row. Once these estimates are provided back to the network, the expected reward can be calculated in the target layer \Target.


\subsection{Competencies (Latent or Explicit)} 
\label{sec:latent-skills}
As with the "avoid wall" latent competency for DoorKey, we propose that there is some shared latent set of competencies of which mastery over can predict an agent's success rate on different metrics in different environments. More concretely, let $\kappa_i \in K$ be a set of observable metrics, which can be any measurable target (e.g., a standard reward function, a shaped or partitioned reward function, or the completion an option). For example, we can define a binary variable that is 1 if an agent received a reward of more than a threshold value, and 0 otherwise. 

Drawing inspiration from ECD, we propose that there exists a set of latent competencies $\Psi$ that are not directly observable but can be inferred from the observed metrics. These competencies are such that the probability of success on a given observable metric $\kappa_i$ of an agent on a specific task $m_i$ is dependent on sufficient mastery of the corresponding latent competencies. This relationship allows us estimate the impact of competencies on unseen tasks. A BN allows us to model this relationship, estimate competency levels from data, and subsequently estimate success rates on different environments. 

Explicit competencies can be derived from techniques that decompose a task into subtasks.  
For example, in hierarchical planning, a method decomposes an abstract task.  
A set of such methods could be used to construct the competencies.
For example, recent work has used hierarchical goal networks to decompose tasks and train RL policies. \cite{patra2022hierarchical,patra2023relating}.
They call the resulting policies goal skills.
For the SEBN defined in Fig. \ref{fig:doorkeysebn}, all four of the competencies in $\Psi$ could be defined as an explicit goal skills.
In the case of this SEBN, the dependencies in the network are exactly the same as a corresponding hierarchical goal skill network. Consequently, we could take any problem with a heirarchical goal skill network, define environment-conditioned dependencies, and turn it into a SEBN. 

The flexibility of letting competencies be latent or explicit allows us to model environments where intermediate decompositions may not be well-defined.
In the absence of an easy way to check if an agent satisfies a goal for a given goal-skill, then it can be set to be a latent competency in the SEBN.

\subsection{Skill Environment Bayesian Network}
We can now define a Skill Environment Bayesian Network (SEBN) for estimating the competency level of an agent and modeling the relationship between agent competency levels, env features, and observable goals/metrics. The SEBN is a tuple $\{X, D, \parentfunctions\}$, where the variable set $X = E \cup \Target \cup \Proficiency$ is split into three sets of variables:

\textbf{Environment Variables.} $E$ is a set of variables that represent the features of an environment descriptor. Each variable in this set corresponds to a specific feature of the environment (and thus the domain of a given variable is the set of possible values the corresponding environment feature can take). In the gridworld example, $E$ consists of features for wall, door, and distance. 

    \textbf{Target Variables.} \Target is set of variables that directly correspond one or more targets. If there is a single policy, then \Target will have a single node, as in Figure~\ref{fig:sebn-bipedal-walker}. But if there are options available to the agent, then each $\target{j} \in K$ corresponds to the option for task $\taski^j$ (cf. \secref{sec:options}). These variables then provide estimates of the value of executing that option in the current environment. For the purposes of this paper, that estimate is thresholded such that each variable returns a boolean value corresponding to whether its estimate meets a performance threshold (roughly corresponding to an estimate of whether the option will succeed or fail). 

    \textbf{Competency Variables.}  \Proficiency contains the set of variables that represent the competency levels of an agent. Each variable in this set denotes the level of proficiency an agent has in a particular competency and takes values in a range from $\{0,1,...,N\}$ where $N$ is the highest proficiency level for a given competency. For this paper, we will define competency with two or three levels of proficiency. Competency levels are roughly ordered, as provided in a set of requirement specifications, by a human expert. The rationale for writing these specifications is to convey whether a given environment requires sufficient proficiency in several competencies. Specifications follow (roughly) ordered values of competency (e.g. a higher “move” competency should enable harder tasks). The competencies can be latent (e.g., capturing whether an agent avoids obstacles while moving) or explicitly learnable (cf. \secref{sec:latent-skills}). 

The parent functions \parentfunctions provide the distribution of possible values of each variable conditioned on their parent variables. To construct these parent functions, we specify a list of competency requirements that is procedurally used to construct the corresponding CPTs. We provide specific detail about this process in \secref{sec:variable-distributions}. 

Once constructed, we can use the SEBN to estimate an agent's competency levels and determine what environments or competencies the agent should focus on learning next. To do this, we must estimate two quantities for a task $\taski \in \tasks$: 
\textbf{Competency Level:} $P(\Proficiency = \proficiency| \target{i} = r, \tasks = \taski)$ - the probability that an agent has a competency level of \proficiency, given its prior performance of at least reward $r$ on target \target{i} for task \taski.  
\textbf{Expected task reward:} $P(\target{i} = r| \tasks = \taski, \Proficiency = \proficiency)$ - the probability that an agent can achieve a reward of at least $r$ for target \target{i} conditioned on task \taski with a given competency level \proficiency.

We will estimate the competency levels 
using past rollouts.
We will then apply the estimates of competency levels to estimate the expected task reward 
over a collection of candidate environments $\taski \in \tasks$. 
These estimated probabilities will be used to determine which environments the agent should train on next.


\subsection{Defining variable distributions} 
\label{sec:variable-distributions}
To complete our network definition, we need to define the distributions of each variable in the network. The distribution of variables in $\Psi$ and \Target are defined using a hierarchical structure. We start by defining the leaves of this structure which are located in \Proficiency. 

\paragraph{Defining \Proficiency}
Consider the "move" competency $move \in \Psi$ in the gridworld navigation environment. In Fig. \ref{fig:doorkeysebn}, we define $move$ as having three levels of proficiency $\{0,1,2\}$, associated with a corresponding parameter set $\phi_{move} = \{0.8, 0.2, 0.0\}$ such that the probability of the "move" at competency level $j$ is given by:
$P(move = j) = \phi_{move}(j)$.
In this case, the parameters denote that currently the agent has a $move$ competency of 80\% probability for no mastery and a 20\% probability of having level one mastery.

More generally, \Proficiency can have a hierarchical structure and there can be latent competencies within $\Psi$ that depend on other latent competencies.  
To allow for these hierarchies in \Proficiency, we define $B$ to be a subset of $\Psi$ which represents a base set of competencies (the leaves). The distribution of these base competencies is determined by a set of associated parameters $\phi_{B}$. 

\paragraph{Defining \Target}
The variables in \Target represent the success of an agent on a given target. In an SEBN, we seek to model the relationship where the success probability of an agent on a variable $\target{i} \in \Target$ is dependent on three types of parent variables:
(1) the environment features $e_i \in E$ relevant to $\target{i}$
(2) the agent's current competency levels $\psi_i \in \Psi$ 
(3) other targets $\target{i}' \in K$
This means that variables in \Target depend on other variables in $\Psi$ and \Target as well as a set of variables in $E$. The success of a task depends on the agent's mastery of the competencies required for a given environment configuration and an independent failure rate $\lambda$. A task succeeds with a rate of $(1-\lambda)$, if all sub-requirements in \Target and $\Psi$ are satisfied. 

More concretely, for each relevant environment configuration $e_i$ in relation to a goal variable $\target{i} \in K$, we define a set of competency level requirements $R_{\target{i}}$ that an agent must master to successfully complete the task of $k_i$ on the environment $e_i$. For example, in Fig. \ref{fig:doorkeysebn}, \textit{haskey} depends on the \textit{distance} environment feature and the latent $move$ and $pickup$ competencies. Suppose we define the following competency level requirements for the \textit{haskey} node:
\begin{small}
\begin{itemize}[noitemsep,topsep=0pt,parsep=0pt,partopsep=0pt]
    \item[] haskey: (distance=0 ~|~ move=1,pick up = 1) 
    \item[] haskey: (distance=1 ~|~ move=2, pick up = 1)
\end{itemize}
\end{small}
These state that if the key is close (distance=0), the agent needs a level of proficiency of 1 in the move and pick up competencies to successfully get the key. However, if the key is far (distance=1), the agent needs a higher level of proficiency in the move competency (move=2). The agent should have a high chance of success if it meets all necessary requirements and a high chance of failure otherwise. 

We directly translate these requirements into entries in the corresponding CPTs in the following way.
    For the first competency level (distance=0), we have that:
    \begin{small}
    \begin{itemize}[noitemsep,topsep=0pt,parsep=0pt,partopsep=0pt]
        \item[] $P(haskey = 1| distance = 0, move = 0, pick up = 0) = \lambda $
        \item[] $P(haskey = 1| distance = 0, move = 0, pick up = 1) = \lambda$ 
        \item[] $P(haskey = 1| distance = 0, move = 1, pick up = 0) = \lambda$ 
        \item[] $P(haskey = 1| distance = 0, move >= 1, pick up >= 1) = (1 - \lambda)$
    \end{itemize}
    \end{small}
    For the next competency level (distance=1), we have that:
    \begin{small}
    \begin{itemize}[noitemsep,topsep=0pt,parsep=0pt,partopsep=0pt]
        \item[] $P(haskey = 1| distance = 1, move = 0, pick up = 0) = \lambda$
        \item[] $P(haskey = 1| distance = 1, move = 0, pick up = 1) = \lambda$
        \item[] $P(haskey = 1| distance = 1, move = 1, pick up = 0) = \lambda$
        \item[] $P(haskey = 1| distance = 1, move = 1, pick up = 1) = \lambda$
        \item[] $P(haskey = 1| distance = 1, move >= 2, pick up >= 1) = (1 - \lambda)$
    \end{itemize}
    \end{small}
These state that the agent has a success rate of $(1-\lambda)$ for a given goal for an environment setting if it satisfies all necessary requirements and a success rate of $\lambda$ if there is one or more requirement missing.

For the environment features set $E$, the distribution of variables in this set is fully controlled for the purpose of curriculum generation. Therefore, the data (samples obtained from rollouts) can be used as the distribution for variables in $E$.

\subsection{Curriculum through Inference}
The general algorithm to generate an SEBN-guided automated curriculum is as follows. First we define a generation of the algorithm as $L$ rollouts. Let $\tasks$ be the set of possible tasks and $P_\tasks(m_i)_t$ be the probability that task $m_i = (e_i, \target{i})$ is chosen in the current generation $t$. We first initialize $P_\tasks(m_i)_0$ to some initial task distribution. This initial weighting can be biased towards easier tasks or can be set to a uniform distribution for better initial competency estimation. Let $\Phi_{B} = \{...,\phi_{B_i},...\}$ be the set of parameters associated with the base competencies $B$ in the SEBN. For each generation, we take the following steps:
\begin{enumerate}[noitemsep,topsep=0pt,parsep=0pt,partopsep=0pt]
    \item Sample $L$ rollouts $(m_i = (e_i, \target{i}) \sim P_\tasks, o_i)$. For each rollout, we record a set of observable metrics $\target{i}$. 
    \item Solve the MLE problem:
    \begin{align}
       \Phi_{B}^* = \text{arg}\max_{\Phi_{B}} \prod_{L} P(E = e_i, K = \target{i}) 
       \label{eq:mle}
    \end{align}
    for $\Phi_{B}^*$, the values of the parameters for the base competencies, which is an estimate of agent proficiency level in those competencies.
    \item Update the task distribution for the next generation: $P_\tasks(m_i) = F(P_{\Phi_{B}^*}(\target{i}|m_i))$ where $F$ is some function that maps the estimated success rate of an environment $P_{\Phi_{B}^*}(\target{i}|m_i)$ to a probability distribution.
\end{enumerate}
For our work, we define $F$ using the following fitness function:
\begin{align}
    F(m_i)_t = (P_{\Phi_{B}^*}(\target{i}|m_i)_t - P_{\Phi_{B}^*}(\target{i}|m_i)_{t-1})^2 \nonumber\\
    P_\tasks(m_i)_{t+1} = 0.5 \cdot \frac{F(m_i)}{\sum_{m_i} F(m_i)} + 0.5 \cdot P_\tasks(m_i)_{t}.
    \label{eq:fitness}
\end{align}
In \cite{stout2010competence}, it was proposed that curricula should focus on problems where the agent improves the most or has the most expected \textit{improvement in competence}. To simulate this in our approach, we choose a fitness function where the fitness of the environment is a function of the difference between the current estimated success rate and the estimated success rate on the last generation's SEBN. We also add a smoothing factor to improve learning stability.

Note that our curriculum is agnostic to the choice of learning algorithm and policy which can be assumed to be black boxes. The curriculum only requires observations of rollouts and not the internal reward structure of a given policy. 

\begin{algorithm}
\SetAlgoLined
\begin{small}
\While{ not converged}{
    Sample L environments $m_i = (e_i,\target{i}) \sim P_{\tasks_0}$ \\
    \For{$i = 1 \rightarrow L$}{
        Collect rollout data $(m_i, \target{i})$ while training policy $\pi$
    }
    Solve the MLE problem (Equation \ref{eq:mle}) for estimated competency level $\Phi_{B}^*$ on SEBN $(X,D,\Phi)$ \\
    For candidate environments, estimate agent success rate $P_{\Phi_{B}^*}(\target{i}|m_i)$  using SEBN with $\Phi$ updated with $\Phi_{B}^*$ \\
    For each candidate environment $m_i$, update $P_\tasks(m_i)_{t+1}$ using expected improvement weighting of $P_{\Phi_{B}^*}(\target{i}|m_i)$ (Equation~\ref{eq:fitness})
}
 \caption{SEBN-guided Automated Curriculum\newline
 \textbf{Input:} Initial tasks $P_{\tasks_0}$, Generation size $L$, SEBN $(X,D,\Phi)$ \newline
 \textbf{Initialize:} Initialize policy $\pi$ }
\end{small}
 \label{alg:SEBN}
\end{algorithm}

\subsection{Candidate Selection}
\label{sec:approximate-inference}

On a domain such as Megagrid, we can evaluate $P_{\Phi_{B}^*}(\target{i}|m_i)$ for every combination of environmental features. However, calculating $P_{\Phi_{B}^*}(\target{i}|m_i)$ for all possible environments in BipedalWalker took too much time at the end of each rollout generation. In general, for domains with a large amount of environmental features, it is impractical to evaluate $P_{\Phi_{B}^*}(\target{i}|m_i)$ for every single task $m_i \in \tasks$.  

One way to solve this computational problem is to search in the space of possible environment configurations and only update the distribution of the most promising environments. For our search procedure, we adapt a greedy sample-search procedure based on KL-Search from \cite{hsiao2024surrogate}. This algorithm employs a heuristic search along variables in a Bayesian network to minimize a KL distance heuristic between two networks. By modifying algorithm to instead search for nodes with maximum differences between the current generation's SEBN and the next generation's SEBN updated with $\Phi_{B}^*$, we can find environments where the estimated success rate changes the most. This modification produces a tree search algorithm that selects nodes in the OR-tree corresponding to a given SEBN where the difference $(P_{\Phi_{s_i}^*}(\target{i}|m_i)_t - P_{\Phi_{s_i}^*}(\target{i}|m_i)_{t-1})$ is greatest. 

Given a partial configuration $X$, we use the following heuristic:
\begin{align}
    h(X_1 {=} 0) = |\log (P_{t-1}(X)) - \log (P_t(X))| \cdot P_t(X)
    \label{eq:heuristic}
\end{align}
where $P_{t}(X_1 {=} 0) = P_{\Phi_{s_i}^*}(\target{i}|X1)_t$. 
We use Weighted Mini-bucket Elimination \cite{liu2011bounding} with an ibound of 20 to estimate SEBN probabilities when exact inference is too computationally expensive. Once we select $N = 20$ nodes on the OR-tree using this method, we perform the same calculations from Eq. \ref{eq:fitness} over the 20 selected environment configurations to define our curriculum for the next generation.

\section{Experimental Evaluation}
To demonstrate the effect of our proposed curriculum learning approach, we evaluate the SEBN curriculum on three environments.
\textbf{DoorKey:} A MiniGrid \cite{MinigridMiniworld23} inspired domain with explicit intermediate goal skills. In this domain, the agent must learn to navigate through a grid-based environment to reach a goal location, while also learning to achieve intermediate goals along the way.
\textbf{BipedalWalkerHardcore:} a simulated bipedal robot must learn to walk forward as quickly as possible while maintaining balance. The bipedal robot can encounter a variety of obstacles such as rough terrain, stumps, pits and stairs that need robust policies.
\textbf{Robosuite:} a robotic arm (Kinova Gen 3) must learn to open a door with different weight and latch settings.

In each of these domains, we compare the performance of reinforcement learning agents trained with and without our proposed SEBN-guided curriculum as well as two additional controls:
    \textbf{Uniform curriculum (or Domain Randomization \cite{tobin2017domain}):} all environments have an equal probability of being selected.
    \textbf{Anti-curriculum:} the probability difference in Eq. \ref{eq:fitness} is replaced with (1 - difference) and we use a min priority queue for candidate selection during the sample-search algorithm.

We evaluate the agents on their ability to learn effective policies that can achieve high rewards in each domain, as well as their ability to generalize to new tasks and environments. Our results show that the SEBN-guided curriculum consistently improves the performance of reinforcement learning agents across all three domains. All runs are performed on an AMD EPYC 7H12 64 core CPU with networks being handled on an A100 GPU.

\begin{figure}
    \includegraphics[width=0.36\textwidth]{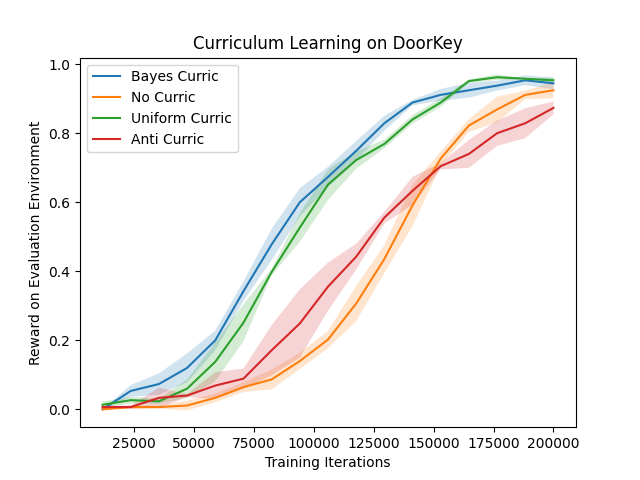}
    \caption{Result of employing a SEBN-guided automated curriculum on the DoorKey environment. 
    }
    \label{fig:doorkeyres}
\end{figure}

\paragraph{DoorKey}
We start with a gridworld environment named Megagrid based on MiniGrid \cite{MinigridMiniworld23}; we reimplemented this standard gridworld environment to enable easier generation of environments using the task descriptor\footnote{Please send an email to the authors to request Megagrid code}. We evaluate on the simple DoorKey environment to evaluate the effectiveness of our curriculum learning approach when combined with explicit goal-skills. We use the SEBN shown in Fig. \ref{fig:doorkeysebn} which has the following variables:
\begin{itemize}[noitemsep,topsep=0pt,parsep=0pt,partopsep=0pt]
    \item[$E$:] includes a wall feature \zeroone, door feature \zeroone, and a distance feature \zeroone. A selection of the different environments that can be experience by an agent can seen in Fig. \ref{fig:doorkeyenv}. 
    \item[$\Psi$:] includes "move to" \zot, "pick up" \zeroone, "avoid wall" \zeroone, "drop" \zeroone, and "open door" \zeroone.
    \item[\Target:] includes three options: "at(goal)" \zeroone, "opened(door)" \zeroone, and "has(key)" \zeroone~ each trained with a PPO policy.
\end{itemize}
Observations of the environment are provided as partially observed cardinal direction data, following the sensor convention for the Lightworld domain in \cite{konidaris2012transfer}. We assume that the agent has four cardinal sensors for each item. For example, in the rightmost env of Fig. \ref{fig:doorkeyenv}, the agent would receive the observation that there is a key one step above it, a wall one step below it, and the goal 4 steps below it (0.875 key - up, 0.875 wall - down, 0.5 goal - down). An agent gets a reward of 1 if it reaches the goal square or completes any intermediate goals required (e.g. picking up a key or opening a door). There is no step penalty but the episode is automatically terminated if no progress has been made in 50 steps.

We train the policies using an Actor-Critic architecture \cite{sutton2018reinforcement} trained using Proximal Policy Optimization (PPO) \cite{schulman2017proximal}. The policy and value networks each have four hidden layers which are used to calculate their corresponding outputs.

The evaluation curve of our policies can be seen in Fig. \ref{fig:doorkeyres}. This evaluation is performed on the hardest environment (rightmost environment in Fig. \ref{fig:doorkeyenv}). Because this environment is relatively simple, it is easy for policies learned without a curriculum to achieve a high success rate. However, there is a noticeable jumpstart where the SEBN-guided curriculum provides a gain in learning efficiency.

We performed a generalization experiment where the learned policies are transferred to an evaluation on an \emph{unseen} and much larger 32 x 32 grid environment. The policies learned using the SEBN-guided curriculum are more robust to this change in grid size succeeded at an average rate of 93 percent compared to only 82 percent without a curriculum. 


\paragraph{BipedalWalker Hardcore}
For the next domain, we evaluate our SEBN-guided curriculum on BipedalWalker (BPW), a continuous control environment. We employ a modified version of the BipedalWalkerHardcore environment from \cite{parker2022evolving} to suit a limited computational budget. We define the following SEBN:
\begin{itemize}[noitemsep,topsep=0pt,parsep=0pt,partopsep=0pt]
    \item[$E$:] includes five design parameters: ground roughness $\{0 - 7\}$, pit gap $\{0 - 3\}$, stump height $\{0 - 2\}$, stair width $\{0 - 3\}$, and stair steps $\{0 - 3\}$. Since there are too many different environments to perform exact inference, it is necessary to use a sample-search procedure to select candidate environments.
    \item[$\Psi$:] includes  "move", "climb", "jump", "balance", "descend". All with proficiency levels $\{0,1,2\}$. 
    \item[\Target:] includes one observable metric \zeroone: whether the agent has traveled a distance of 30 units ($\approx$1/3rd of the level).
\end{itemize}
The observation of the agent consists of internal sensor measurements such as (hull angle speed, angular velocity, horizontal velocity, etc..) an a set of 10 lidar rangefinder measurements. On this environment, the robotic walker gets a dense positive reward for traveling forward on the terrain, a small negative reward for using its motors, and a negative reward of -100 if it falls down. On environments that are too challenging for the agent at its current capabilities, this reward structure can promote a locally optimal behavior of simply staying still and preventing itself from falling. 

To learn our policies, we use the TD3-Fork algorithm from Honghao et al. \cite{wei2020fork}, which was shown to train much faster than standard PPO on BPW. We train agents for 3.5 million timesteps. During training we evaluate against four specific challenges (Stairs, PitGap, Stump, and Roughness in Fig. \ref{fig:bipedalwalkerenv}) as well as a combined environment (Evaluation) that contains all challenges together. We compare against a policy trained on only the combined environment.

\begin{figure}
    \includegraphics[width=0.36\textwidth]{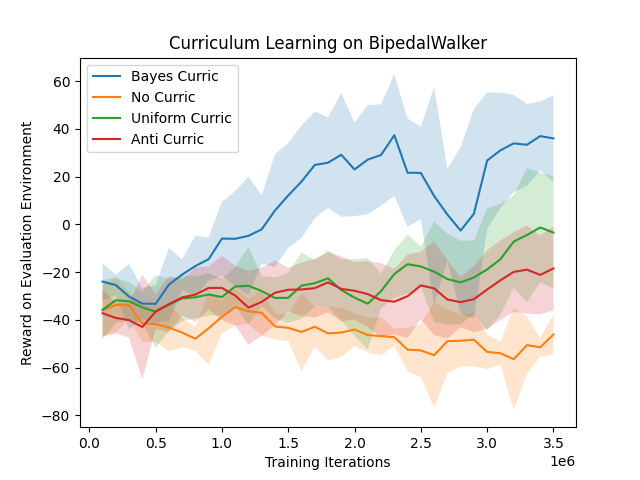}
    \caption{SEBN-guided automated curriculum on BipedalWalker. Evaluation environments are randomly generated within a given environment feature set.
    }
    \label{fig:bipedalwalkerres}
\end{figure}
The results of the evaluation on the combined environment can be seen in Fig. \ref{fig:bipedalwalkerres}. Since we do not train for a large number of environment timesteps, we can observe that without a curriculum, TD3-Fork does not manage to learn to the point of a positive reward on any of the test environments.

There is a significant reward divergence in the results starting at 1 million timesteps. The uniform and anti curriculum perform better than having no curriculum with the uniform curriculum perform marginally better than the anti-curriculum. It can be seen on each graph that the policies trained using the SEBN-guided curriculum manages to get to a point where the agent starts receiving a positive reward for each environment, getting past the initial hurdle of the locally optimal staying still behavior.

It is interesting to observe that the learning curve for the SEBN-guided curriculum has much higher variance than the learning curves for other methods. Due to the size of the environment design space, it is necessary to use approximate inference techniques to search for candidate environments. Because we use a sample-search procedure, the search process is not guaranteed to find environments with the largest heuristic difference (see Eq. \ref{eq:heuristic}). This means that sub-optimal environments can be introduced into the curriculum. This introduces a large factor into learning curve variance beyond standard noise from reinforcement learning algorithms.


\begin{figure}
    \includegraphics[width=0.36\textwidth]{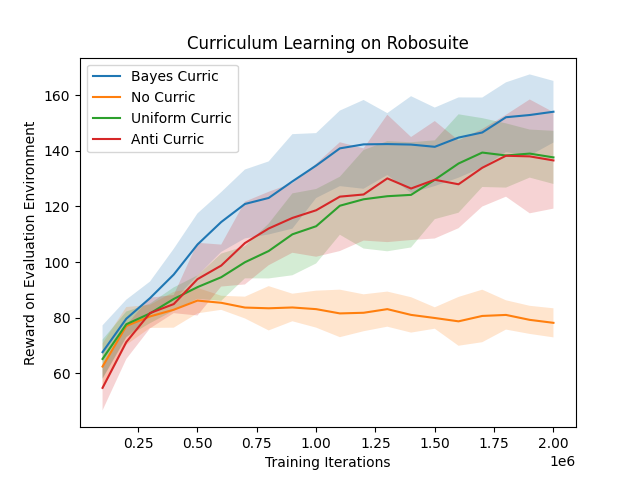}
    \caption{SEBN-guided automated curriculum on the Robosuite Door environment. 15 policies are learned for each line and evaluated on the hardest (mass=6, latch=1) environment.}
    \label{fig:robosuiteres}
\end{figure}

\paragraph{Robosuite - Open Door}
In our final domain, we evaluate the SEBN-guided curriculum in a simulated robotics domain using the robosuite simulation environment. In particular we choose the Door task in which an agent needs to manipulate a robot to open a door. For our environment design space, we include two main parameters: the weight of the door (with 6 different settings) and whether the door has a latch or not. The specific robot that we choose to simulate is a Kinova3 arm. 
We define the following SEBN:
\begin{itemize}[noitemsep,topsep=0pt,parsep=0pt,partopsep=0pt]
    \item[$E$:] includes design parameters mass $\{0 -6 \}$ and latch \zeroone.
    \item[$\Psi$:] includes "move arm" \zeroone, "unlock" \zeroone, "door" \zot. 
    \item[\Target:] includes one observable metric \zeroone: whether the agent has successfully opened the door.
\end{itemize}

To learn our policies, we use PPO on a neural network with two hidden layers. We use the inbuilt observation setup and reward shaping in the robosuite environment to accelerate the learning process and we train our agents for 2 million timesteps. We evaluate our learned policies every 100,000 timesteps and the results of the experiment can be seen in Fig. \ref{fig:robosuiteres}. 

It is interesting to observe that the default training method stagnates after reaching a reward plateau. When viewing the actual behavior of the learned policy, this reward plateau is indicative of learning a policy of moving the arm towards the door handle but not actually moving the door. It is possible that either the weight of the door or the presence of a latch in harder environments prevents the agent from attempting the difficult action of applying force to the door to get an increased reward. Since the SEBN-guided curriculum will have started with at least some of its distribution in the easy case of a light door with no latch, the agent will have learned that opening the door can give a positive reward and transfer this behavior to more difficult environments. 

We also performed a generalization study, where we test our learned policies on an \emph{unseen} heavy door env not included during our learning process. We observed that with the policy learned by the SEBN-guided curriculum is easily transferred to more heavy doors (obtaining a reward of 240 on the heavy door env compared to 150 without a curriculum), in contrast to the other methods. 

\section{Related Work} 
\paragraph{Automated Curriculum Generation}  Several topics relate to ordering tasks to improve learning performance. A few approaches have considered the problem of estimating agent skill competencies. In the context of education, in addition to ECD \cite{mislevy2003brief}, another approach close to ours is that of \citeauthor{greenEtal2011.iaai.learningASkillTeachingCurriculum} \cite{{greenEtal2011.iaai.learningASkillTeachingCurriculum}}, who used a BN to determine the next task for the human student.  
This approach is similar to \cite{kumar2024practice}, which considered an active learning problem in a robotics domain of choosing which skills to practice to maximize future task success, which involves estimating the competence of each skill and situating it in the task distribution through competence-aware planning. In contrast to our approach, they employ a simplified Bayesian time series model that does not relate environmental features with goal and skill competencies. This limits the applicability of their approach towards only choosing what skill to train and not the agent's environment. Similar to our selection process, \citeauthor{BalleraEtal2014.icetc.personalizingElearning} \cite{BalleraEtal2014.icetc.personalizingElearning} used a roulette wheel selection of tasks. 

A related area is the literature on Unsupervised Environment Design (UED) \cite{dennis2020emergent} and other developed mechanisms for curating environments based on a regret heuristic \cite{jiang2021replay}. In prior UED approaches such as PAIRED \cite{dennis2020emergent}, the agent's curriculum is generated using a regret-based heuristic. The heuristic is typically an estimate of the true regret, since the optimal policy is unknown. In PAIRED, this heuristic is calculated by learning an antagonistic policy and evaluating the difference between it and the protagonist policy. In contrast, like ACCEL \cite{parker2022evolving}, our method does not need to learn a second antagonistic policy and instead uses rollouts from a single agent to compute the next part of the curriculum. In contrast to ACCEL, 
our curriculum does not rely on local changes and can incorporate larger jumps in environment selection. Furthermore, while it is necessary to use rollouts on each environment for ACCEL to obtain a regret estimate, we can estimate success rates on unseen environments by leveraging the relationships encoded between competencies and environmental features in our SEBN.

\paragraph{Task Descriptors}
As mentioned in \secref{sec:taskdescriptors}, grouping tasks using features, i.e., task descriptors, are a well understood technique for task creation (\cite{rostamiEtAl20.jair.usingTaskDescriptions,isele2016task,narvekar2016source,konidaris2012transfer}).  The key idea in these works is to facilitate learning transfer by creating similar tasks that share common features.  These features can leave certain variables free during task construction that enable a family of similar tasks. 
Our work supplements prior work by adding the target of the task to the task descriptor, allowing the curriculum to emphasize subtasks.

To our knowledge, there has been limited previous work in integrating curriculum learning on both skills and environment features. However, it can be said that our research expands on the concept of using task descriptors in the creation of automated curricula. In \cite{patra2023relating}, a task-graph curricula is used to generate a curriculum over tasks and environmental features. However, they employ a simple greedy best-first search on the task-graph to choose an order for their curriculum. This is different from our approach that updates a distribution over the task-graph and dynamically adjusts this distribution based on data from rollouts.

\paragraph{Hierarchical Goal Networks}

The structure of our Skill Environment Bayesian network shares similarity to both goal skill networks and fault diagnosis networks. In fault diagnosis networks \cite{cai2017bayesian}, BNs are used to model the relationship between a set of sensors and a set of faults. In our case, the sensors are analogous to an SEBN's observable goal metrics, and the faults are analogous to an SEBN's skills. An SEBN can then be seen as a fault diagnosis network where different roll-outs are independent tests that determine what latent competencies may have not been mastered.

\paragraph{Expert guidance for RL training}
One of the limitations of the SEBN is its reliance on the expert-provided competencies. 
As noted, these could be derived from hierarchical approaches. 
But providing domain knowledge is common in many hierarchical RL settings.
Similar to our work,  \citeauthor{patra2022hierarchical} \cite{patra2022hierarchical}, provided a hierarchical learning structure.
This kind of expert knowledge is common in  imitation learning (e.g., \cite{zhang19leveraging} \cite{hussein2017imitation} \cite{le2018hierarchical}), where an expert human guides a learning agent.
Providing expert guidance is also common in Hierarchical RL approaches (e.g., \cite{ahmadiTaylorStone07.aamas.IFSA}) and in standard RL approaches (e.g., \cite{andreas2017modular}).
Finally, expert guidance was shown to be helpful for a sparse-reward task in an Object Oriented MDP setting \cite{abelEtAl15.icaps.goalBasedActionPriors}.




\section{Conclusion and Future Work}
We presented a novel method for generating curriculum over environmental features using a Skill-Environment Bayesian Network. This network is used to estimate agent competency level based off of past rollouts and can be used to infer estimated agent success rates on unseen environments. We demonstrate the effectiveness of this approach on a variety of domains.

For this work, we relied on a pre-defined set of skills and environmental features. In future work, we would like to extend the model to handle more open ended environments where we can add new environmental features or agent skills dynamically to the SEBN. It might be interesting to see if we can apply techniques such as GO-MTL \cite{kumar2012learning} for learning a latent space over tasks and approaches for detecting critical regions \cite{molina2020learn} to learn new skills.

There has also been a great deal of interest in the use of Large Language Models (LLMs) in planning domains for the purpose of automatically generating planning models. As SEBNs can be built from a heirarchical goal network, it might stand to reason that LLMs could also be used to automatically generate SEBNs from domain documents. Since there is much more leeway in the skills required (since our method supports latent competency nodes), it may be easier to generate these SEBNs than equivalent goal networks.

\begin{acks}
We thank the Basic Research Office of OUSD and NRL for funding this research.
\end{acks}



\newpage
\balance
\bibliographystyle{ACM-Reference-Format} 
\bibliography{sample}


\begin{thebibliography}{37}


\ifx \showCODEN    \undefined \def \showCODEN     #1{\unskip}     \fi
\ifx \showDOI      \undefined \def \showDOI       #1{#1}\fi
\ifx \showISBNx    \undefined \def \showISBNx     #1{\unskip}     \fi
\ifx \showISBNxiii \undefined \def \showISBNxiii  #1{\unskip}     \fi
\ifx \showISSN     \undefined \def \showISSN      #1{\unskip}     \fi
\ifx \showLCCN     \undefined \def \showLCCN      #1{\unskip}     \fi
\ifx \shownote     \undefined \def \shownote      #1{#1}          \fi
\ifx \showarticletitle \undefined \def \showarticletitle #1{#1}   \fi
\ifx \showURL      \undefined \def \showURL       {\relax}        \fi
\providecommand\bibfield[2]{#2}
\providecommand\bibinfo[2]{#2}
\providecommand\natexlab[1]{#1}
\providecommand\showeprint[2][]{arXiv:#2}

\bibitem[\protect\citeauthoryear{Abel, Hershkowitz, Barth-Maron, Brawner, O'Farrell, MacGlashan, and Tellex}{Abel et~al\mbox{.}}{2015}]%
        {abelEtAl15.icaps.goalBasedActionPriors}
\bibfield{author}{\bibinfo{person}{David Abel}, \bibinfo{person}{David Hershkowitz}, \bibinfo{person}{Gabriel Barth-Maron}, \bibinfo{person}{Stephen Brawner}, \bibinfo{person}{Kevin O'Farrell}, \bibinfo{person}{James MacGlashan}, {and} \bibinfo{person}{Stefanie Tellex}.} \bibinfo{year}{2015}\natexlab{}.
\newblock \showarticletitle{Goal-based action priors}. In \bibinfo{booktitle}{\emph{Proceedings of the International Conference on Automated Planning and Scheduling}}, Vol.~\bibinfo{volume}{25}. \bibinfo{pages}{306--314}.
\newblock


\bibitem[\protect\citeauthoryear{Ahmadi, Taylor, and Stone}{Ahmadi et~al\mbox{.}}{2007}]%
        {ahmadiTaylorStone07.aamas.IFSA}
\bibfield{author}{\bibinfo{person}{Mazda Ahmadi}, \bibinfo{person}{Matthew~E. Taylor}, {and} \bibinfo{person}{Peter Stone}.} \bibinfo{year}{2007}\natexlab{}.
\newblock \showarticletitle{IFSA: incremental feature-set augmentation for reinforcement learning tasks}. In \bibinfo{booktitle}{\emph{Proceedings of the 6th international joint conference on Autonomous agents and multiagent systems}} \emph{(\bibinfo{series}{AAMAS ’07})}. \bibinfo{publisher}{Association for Computing Machinery}, \bibinfo{address}{New York, NY, USA}, \bibinfo{pages}{1–8}.
\newblock
\showISBNx{978-81-904262-7-5}
\urldef\tempurl%
\url{https://doi.org/10.1145/1329125.1329351}
\showDOI{\tempurl}


\bibitem[\protect\citeauthoryear{Andreas, Klein, and Levine}{Andreas et~al\mbox{.}}{2017}]%
        {andreas2017modular}
\bibfield{author}{\bibinfo{person}{Jacob Andreas}, \bibinfo{person}{Dan Klein}, {and} \bibinfo{person}{Sergey Levine}.} \bibinfo{year}{2017}\natexlab{}.
\newblock \showarticletitle{Modular multitask reinforcement learning with policy sketches}. In \bibinfo{booktitle}{\emph{International conference on machine learning}}. PMLR, \bibinfo{pages}{166--175}.
\newblock


\bibitem[\protect\citeauthoryear{Ballera, Lukandu, and Radwan}{Ballera et~al\mbox{.}}{2014}]%
        {BalleraEtal2014.icetc.personalizingElearning}
\bibfield{author}{\bibinfo{person}{Melvin Ballera}, \bibinfo{person}{Ismail~Ateya Lukandu}, {and} \bibinfo{person}{Abdalla Radwan}.} \bibinfo{year}{2014}\natexlab{}.
\newblock \showarticletitle{Personalizing E-learning curriculum using: reversed roulette wheel selection algorithm}. In \bibinfo{booktitle}{\emph{2014 International Conference on Education Technologies and Computers (ICETC)}}. \bibinfo{pages}{91–97}.
\newblock
\showISSN{2155-1812}
\urldef\tempurl%
\url{https://doi.org/10.1109/ICETC.2014.6998908}
\showDOI{\tempurl}


\bibitem[\protect\citeauthoryear{Bengio, Louradour, Collobert, and Weston}{Bengio et~al\mbox{.}}{2009}]%
        {bengio2009curriculum}
\bibfield{author}{\bibinfo{person}{Yoshua Bengio}, \bibinfo{person}{J{\'e}r{\^o}me Louradour}, \bibinfo{person}{Ronan Collobert}, {and} \bibinfo{person}{Jason Weston}.} \bibinfo{year}{2009}\natexlab{}.
\newblock \showarticletitle{Curriculum learning}. In \bibinfo{booktitle}{\emph{Proceedings of the 26th annual international conference on machine learning}}. \bibinfo{pages}{41--48}.
\newblock


\bibitem[\protect\citeauthoryear{Cai, Huang, and Xie}{Cai et~al\mbox{.}}{2017}]%
        {cai2017bayesian}
\bibfield{author}{\bibinfo{person}{Baoping Cai}, \bibinfo{person}{Lei Huang}, {and} \bibinfo{person}{Min Xie}.} \bibinfo{year}{2017}\natexlab{}.
\newblock \showarticletitle{Bayesian networks in fault diagnosis}.
\newblock \bibinfo{journal}{\emph{IEEE Transactions on industrial informatics}} \bibinfo{volume}{13}, \bibinfo{number}{5} (\bibinfo{year}{2017}), \bibinfo{pages}{2227--2240}.
\newblock


\bibitem[\protect\citeauthoryear{Darwiche}{Darwiche}{2009}]%
        {darwiche2009modeling}
\bibfield{author}{\bibinfo{person}{Adnan Darwiche}.} \bibinfo{year}{2009}\natexlab{}.
\newblock \bibinfo{booktitle}{\emph{Modeling and reasoning with Bayesian networks}}.
\newblock \bibinfo{publisher}{Cambridge university press}.
\newblock


\bibitem[\protect\citeauthoryear{Dechter}{Dechter}{2013}]%
        {dechter2013reasoning}
\bibfield{author}{\bibinfo{person}{Rina Dechter}.} \bibinfo{year}{2013}\natexlab{}.
\newblock \showarticletitle{Reasoning with probabilistic and deterministic graphical models: Exact algorithms}.
\newblock \bibinfo{journal}{\emph{Synthesis Lectures on Artificial Intelligence and Machine Learning}} \bibinfo{volume}{7}, \bibinfo{number}{3} (\bibinfo{year}{2013}), \bibinfo{pages}{1--191}.
\newblock


\bibitem[\protect\citeauthoryear{Dennis~et al.}{Dennis~et al.}{2020}]%
        {dennis2020emergent}
\bibfield{author}{\bibinfo{person}{Michael Dennis~et al.}} \bibinfo{year}{2020}\natexlab{}.
\newblock \showarticletitle{Emergent complexity and zero-shot transfer via unsupervised environment design}.
\newblock \bibinfo{journal}{\emph{Advances in neural information processing systems}}  \bibinfo{volume}{33} (\bibinfo{year}{2020}), \bibinfo{pages}{13049--13061}.
\newblock


\bibitem[\protect\citeauthoryear{et~al.}{et~al.}{2023}]%
        {MinigridMiniworld23}
\bibfield{author}{\bibinfo{person}{Maxime Chevalier-Boisvert et al.}} \bibinfo{year}{2023}\natexlab{}.
\newblock \showarticletitle{Minigrid \& Miniworld: Modular \& Customizable Reinforcement Learning Environments for Goal-Oriented Tasks}.
\newblock \bibinfo{journal}{\emph{CoRR}}  \bibinfo{volume}{abs/2306.13831} (\bibinfo{year}{2023}).
\newblock


\bibitem[\protect\citeauthoryear{Green, Walsh, Cohen, and Chang}{Green et~al\mbox{.}}{2011}]%
        {greenEtal2011.iaai.learningASkillTeachingCurriculum}
\bibfield{author}{\bibinfo{person}{Derek Green}, \bibinfo{person}{Thomas Walsh}, \bibinfo{person}{Paul Cohen}, {and} \bibinfo{person}{Yu-Han Chang}.} \bibinfo{year}{2011}\natexlab{}.
\newblock \showarticletitle{Learning a Skill-Teaching Curriculum with Dynamic Bayes Nets}.
\newblock \bibinfo{journal}{\emph{Proceedings of the AAAI Conference on Artificial Intelligence}} \bibinfo{volume}{25}, \bibinfo{number}{2} (\bibinfo{date}{Aug.} \bibinfo{year}{2011}), \bibinfo{pages}{1648–1654}.
\newblock
\showISSN{2374-3468, 2159-5399}
\urldef\tempurl%
\url{https://doi.org/10.1609/aaai.v25i2.18855}
\showDOI{\tempurl}


\bibitem[\protect\citeauthoryear{Hsiao, Nau, Pezeshki, and Dechter}{Hsiao et~al\mbox{.}}{2024}]%
        {hsiao2024surrogate}
\bibfield{author}{\bibinfo{person}{Vincent Hsiao}, \bibinfo{person}{Dana~S Nau}, \bibinfo{person}{Bobak Pezeshki}, {and} \bibinfo{person}{Rina Dechter}.} \bibinfo{year}{2024}\natexlab{}.
\newblock \showarticletitle{Surrogate Bayesian Networks for Approximating Evolutionary Games}. In \bibinfo{booktitle}{\emph{International Conference on Artificial Intelligence and Statistics}}. PMLR, \bibinfo{pages}{2566--2574}.
\newblock


\bibitem[\protect\citeauthoryear{Hussein, Gaber, Elyan, and Jayne}{Hussein et~al\mbox{.}}{2017}]%
        {hussein2017imitation}
\bibfield{author}{\bibinfo{person}{Ahmed Hussein}, \bibinfo{person}{Mohamed~Medhat Gaber}, \bibinfo{person}{Eyad Elyan}, {and} \bibinfo{person}{Chrisina Jayne}.} \bibinfo{year}{2017}\natexlab{}.
\newblock \showarticletitle{Imitation learning: A survey of learning methods}.
\newblock \bibinfo{journal}{\emph{ACM Computing Surveys (CSUR)}} \bibinfo{volume}{50}, \bibinfo{number}{2} (\bibinfo{year}{2017}), \bibinfo{pages}{1--35}.
\newblock


\bibitem[\protect\citeauthoryear{Isele, Rostami, and Eaton}{Isele et~al\mbox{.}}{2016}]%
        {isele2016task}
\bibfield{author}{\bibinfo{person}{David Isele}, \bibinfo{person}{Mohammad Rostami}, {and} \bibinfo{person}{Eric Eaton}.} \bibinfo{year}{2016}\natexlab{}.
\newblock \showarticletitle{Using task features for zero-shot knowledge transfer in lifelong learning}.
\newblock \bibinfo{journal}{\emph{Proc. {IJCAI}}} (\bibinfo{year}{2016}).
\newblock


\bibitem[\protect\citeauthoryear{Jiang~et al.}{Jiang~et al.}{2021}]%
        {jiang2021replay}
\bibfield{author}{\bibinfo{person}{Minqi Jiang~et al.}} \bibinfo{year}{2021}\natexlab{}.
\newblock \showarticletitle{Replay-guided adversarial environment design}.
\newblock \bibinfo{journal}{\emph{Advances in Neural Information Processing Systems}}  \bibinfo{volume}{34} (\bibinfo{year}{2021}), \bibinfo{pages}{1884--1897}.
\newblock


\bibitem[\protect\citeauthoryear{Konidaris, Scheidwasser, and Barto}{Konidaris et~al\mbox{.}}{2012}]%
        {konidaris2012transfer}
\bibfield{author}{\bibinfo{person}{George Konidaris}, \bibinfo{person}{Ilya Scheidwasser}, {and} \bibinfo{person}{Andrew~G Barto}.} \bibinfo{year}{2012}\natexlab{}.
\newblock \showarticletitle{Transfer in reinforcement learning via shared features}.
\newblock \bibinfo{journal}{\emph{Journal of Machine Learning Research}} (\bibinfo{year}{2012}).
\newblock


\bibitem[\protect\citeauthoryear{Kumar and Daume~III}{Kumar and Daume~III}{2012}]%
        {kumar2012learning}
\bibfield{author}{\bibinfo{person}{Abhishek Kumar} {and} \bibinfo{person}{Hal Daume~III}.} \bibinfo{year}{2012}\natexlab{}.
\newblock \showarticletitle{Learning task grouping and overlap in multi-task learning}.
\newblock \bibinfo{journal}{\emph{arXiv preprint arXiv:1206.6417}} (\bibinfo{year}{2012}).
\newblock


\bibitem[\protect\citeauthoryear{Kumar, Silver, McClinton, Zhao, Proulx, Lozano-P{\'e}rez, Kaelbling, and Barry}{Kumar et~al\mbox{.}}{2024}]%
        {kumar2024practice}
\bibfield{author}{\bibinfo{person}{Nishanth Kumar}, \bibinfo{person}{Tom Silver}, \bibinfo{person}{Willie McClinton}, \bibinfo{person}{Linfeng Zhao}, \bibinfo{person}{Stephen Proulx}, \bibinfo{person}{Tom{\'a}s Lozano-P{\'e}rez}, \bibinfo{person}{Leslie~Pack Kaelbling}, {and} \bibinfo{person}{Jennifer Barry}.} \bibinfo{year}{2024}\natexlab{}.
\newblock \showarticletitle{Practice Makes Perfect: Planning to Learn Skill Parameter Policies}.
\newblock \bibinfo{journal}{\emph{arXiv preprint arXiv:2402.15025}} (\bibinfo{year}{2024}).
\newblock


\bibitem[\protect\citeauthoryear{Le, Jiang, Agarwal, Dud{\'\i}k, Yue, and Daum{\'e}~III}{Le et~al\mbox{.}}{2018}]%
        {le2018hierarchical}
\bibfield{author}{\bibinfo{person}{Hoang Le}, \bibinfo{person}{Nan Jiang}, \bibinfo{person}{Alekh Agarwal}, \bibinfo{person}{Miroslav Dud{\'\i}k}, \bibinfo{person}{Yisong Yue}, {and} \bibinfo{person}{Hal Daum{\'e}~III}.} \bibinfo{year}{2018}\natexlab{}.
\newblock \showarticletitle{Hierarchical imitation and reinforcement learning}. In \bibinfo{booktitle}{\emph{International conference on machine learning}}. PMLR, \bibinfo{pages}{2917--2926}.
\newblock


\bibitem[\protect\citeauthoryear{Liu and Ihler}{Liu and Ihler}{2011}]%
        {liu2011bounding}
\bibfield{author}{\bibinfo{person}{Qiang Liu} {and} \bibinfo{person}{Alexander~T Ihler}.} \bibinfo{year}{2011}\natexlab{}.
\newblock \showarticletitle{Bounding the partition function using holder's inequality}. In \bibinfo{booktitle}{\emph{ICML}}.
\newblock


\bibitem[\protect\citeauthoryear{Mislevy, Almond, and Lukas}{Mislevy et~al\mbox{.}}{2003}]%
        {mislevy2003brief}
\bibfield{author}{\bibinfo{person}{Robert~J Mislevy}, \bibinfo{person}{Russell~G Almond}, {and} \bibinfo{person}{Janice~F Lukas}.} \bibinfo{year}{2003}\natexlab{}.
\newblock \showarticletitle{A brief introduction to evidence-centered design}.
\newblock \bibinfo{journal}{\emph{ETS Research Report Series}} \bibinfo{volume}{2003}, \bibinfo{number}{1} (\bibinfo{year}{2003}), \bibinfo{pages}{i--29}.
\newblock


\bibitem[\protect\citeauthoryear{Molina, Kumar, and Srivastava}{Molina et~al\mbox{.}}{2020}]%
        {molina2020learn}
\bibfield{author}{\bibinfo{person}{Daniel Molina}, \bibinfo{person}{Kislay Kumar}, {and} \bibinfo{person}{Siddharth Srivastava}.} \bibinfo{year}{2020}\natexlab{}.
\newblock \showarticletitle{Learn and link: Learning critical regions for efficient planning}. In \bibinfo{booktitle}{\emph{2020 IEEE International Conference on Robotics and Automation (ICRA)}}. IEEE, \bibinfo{pages}{10605--10611}.
\newblock


\bibitem[\protect\citeauthoryear{Narvekar, Peng, Leonetti, Sinapov, Taylor, and Stone}{Narvekar et~al\mbox{.}}{2020}]%
        {narvekarEtAl20.jmlr.clForRL}
\bibfield{author}{\bibinfo{person}{Sanmit Narvekar}, \bibinfo{person}{Bei Peng}, \bibinfo{person}{Matteo Leonetti}, \bibinfo{person}{Jivko Sinapov}, \bibinfo{person}{Matthew~E. Taylor}, {and} \bibinfo{person}{Peter Stone}.} \bibinfo{year}{2020}\natexlab{}.
\newblock \showarticletitle{Curriculum learning for reinforcement learning domains: A framework and survey}.
\newblock \bibinfo{journal}{\emph{Journal of Machine Learning Research}} \bibinfo{volume}{21}, \bibinfo{number}{181} (\bibinfo{year}{2020}), \bibinfo{pages}{1–50}.
\newblock
\newblock
\shownote{Citation Key: JMLR:v21:20-212.}


\bibitem[\protect\citeauthoryear{Narvekar, Sinapov, Leonetti, and Stone}{Narvekar et~al\mbox{.}}{2016}]%
        {narvekar2016source}
\bibfield{author}{\bibinfo{person}{Sanmit Narvekar}, \bibinfo{person}{Jivko Sinapov}, \bibinfo{person}{Matteo Leonetti}, {and} \bibinfo{person}{Peter Stone}.} \bibinfo{year}{2016}\natexlab{}.
\newblock \showarticletitle{Source task creation for curriculum learning}. In \bibinfo{booktitle}{\emph{Proceedings of the 2016 international conference on autonomous agents \& multiagent systems}}. \bibinfo{pages}{566--574}.
\newblock


\bibitem[\protect\citeauthoryear{Parker-Holder~et al.}{Parker-Holder~et al.}{2022}]%
        {parker2022evolving}
\bibfield{author}{\bibinfo{person}{Jack Parker-Holder~et al.}} \bibinfo{year}{2022}\natexlab{}.
\newblock \showarticletitle{Evolving curricula with regret-based environment design}. In \bibinfo{booktitle}{\emph{International Conference on Machine Learning}}. PMLR, \bibinfo{pages}{17473--17498}.
\newblock


\bibitem[\protect\citeauthoryear{Patra, Cavolowsky, Kulaksizoglu, Li, Hiatt, Roberts, and Nau}{Patra et~al\mbox{.}}{2022}]%
        {patra2022hierarchical}
\bibfield{author}{\bibinfo{person}{Sunandita Patra}, \bibinfo{person}{Mark Cavolowsky}, \bibinfo{person}{Onur Kulaksizoglu}, \bibinfo{person}{Ruoxi Li}, \bibinfo{person}{Laura Hiatt}, \bibinfo{person}{Mark Roberts}, {and} \bibinfo{person}{Dana Nau}.} \bibinfo{year}{2022}\natexlab{}.
\newblock \showarticletitle{A hierarchical goal-biased curriculum for training reinforcement learning}. In \bibinfo{booktitle}{\emph{The international FLAIRS conference proceedings}}, Vol.~\bibinfo{volume}{35}.
\newblock


\bibitem[\protect\citeauthoryear{Patra, Rademacher, Jacobson, Hassold, Kulaksizoglu, Hiatt, Roberts, and Nau}{Patra et~al\mbox{.}}{2023}]%
        {patra2023relating}
\bibfield{author}{\bibinfo{person}{Sunandita Patra}, \bibinfo{person}{Paul Rademacher}, \bibinfo{person}{Kristen Jacobson}, \bibinfo{person}{Kyle Hassold}, \bibinfo{person}{Onur Kulaksizoglu}, \bibinfo{person}{Laura Hiatt}, \bibinfo{person}{Mark Roberts}, {and} \bibinfo{person}{Dana Nau}.} \bibinfo{year}{2023}\natexlab{}.
\newblock \showarticletitle{Relating Goal and Environmental Complexity for Improved Task Transfer: Initial Results}. In \bibinfo{booktitle}{\emph{NeurIPS 2023 Workshop on Generalization in Planning}}.
\newblock


\bibitem[\protect\citeauthoryear{Pearl}{Pearl}{1988}]%
        {pearl}
\bibfield{author}{\bibinfo{person}{J. Pearl}.} \bibinfo{year}{1988}\natexlab{}.
\newblock \bibinfo{booktitle}{\emph{{Probabilistic Reasoning in Intelligent Systems: Networks of Plausible Inference}}}.
\newblock \bibinfo{publisher}{Morgan Kaufmann}.
\newblock


\bibitem[\protect\citeauthoryear{Rostami, Isele, and Eaton}{Rostami et~al\mbox{.}}{2020}]%
        {rostamiEtAl20.jair.usingTaskDescriptions}
\bibfield{author}{\bibinfo{person}{Mohammad Rostami}, \bibinfo{person}{David Isele}, {and} \bibinfo{person}{Eric Eaton}.} \bibinfo{year}{2020}\natexlab{}.
\newblock \showarticletitle{Using Task Descriptions in Lifelong Machine Learning for Improved Performance and Zero-Shot Transfer}.
\newblock \bibinfo{journal}{\emph{{JAIR}}}  \bibinfo{volume}{67} (\bibinfo{year}{2020}), \bibinfo{pages}{673--703}.
\newblock


\bibitem[\protect\citeauthoryear{Schulman, Wolski, Dhariwal, Radford, and Klimov}{Schulman et~al\mbox{.}}{2017}]%
        {schulman2017proximal}
\bibfield{author}{\bibinfo{person}{John Schulman}, \bibinfo{person}{Filip Wolski}, \bibinfo{person}{Prafulla Dhariwal}, \bibinfo{person}{Alec Radford}, {and} \bibinfo{person}{Oleg Klimov}.} \bibinfo{year}{2017}\natexlab{}.
\newblock \showarticletitle{Proximal policy optimization algorithms}.
\newblock \bibinfo{journal}{\emph{arXiv preprint arXiv:1707.06347}} (\bibinfo{year}{2017}).
\newblock


\bibitem[\protect\citeauthoryear{Stout and Barto}{Stout and Barto}{2010}]%
        {stout2010competence}
\bibfield{author}{\bibinfo{person}{Andrew Stout} {and} \bibinfo{person}{Andrew~G Barto}.} \bibinfo{year}{2010}\natexlab{}.
\newblock \showarticletitle{Competence progress intrinsic motivation}. In \bibinfo{booktitle}{\emph{2010 IEEE 9th International Conference on Development and Learning}}. IEEE, \bibinfo{pages}{257--262}.
\newblock


\bibitem[\protect\citeauthoryear{Sutton}{Sutton}{2018}]%
        {sutton2018reinforcement}
\bibfield{author}{\bibinfo{person}{Richard~S Sutton}.} \bibinfo{year}{2018}\natexlab{}.
\newblock \showarticletitle{Reinforcement learning: An introduction}.
\newblock \bibinfo{journal}{\emph{A Bradford Book}} (\bibinfo{year}{2018}).
\newblock


\bibitem[\protect\citeauthoryear{Sutton, Precup, and Singh}{Sutton et~al\mbox{.}}{1999}]%
        {suttonEtAl99.aij.betweenMDPsAndSemiMDPs}
\bibfield{author}{\bibinfo{person}{Richard~S. Sutton}, \bibinfo{person}{Doina Precup}, {and} \bibinfo{person}{Satinder Singh}.} \bibinfo{year}{1999}\natexlab{}.
\newblock \showarticletitle{Between MDPs and semi-MDPs: A framework for temporal abstraction in reinforcement learning}.
\newblock \bibinfo{journal}{\emph{AIJ}} \bibinfo{volume}{112}, \bibinfo{number}{1} (\bibinfo{year}{1999}), \bibinfo{pages}{181--211}.
\newblock


\bibitem[\protect\citeauthoryear{Tobin, Fong, Ray, Schneider, Zaremba, and Abbeel}{Tobin et~al\mbox{.}}{2017}]%
        {tobin2017domain}
\bibfield{author}{\bibinfo{person}{Josh Tobin}, \bibinfo{person}{Rachel Fong}, \bibinfo{person}{Alex Ray}, \bibinfo{person}{Jonas Schneider}, \bibinfo{person}{Wojciech Zaremba}, {and} \bibinfo{person}{Pieter Abbeel}.} \bibinfo{year}{2017}\natexlab{}.
\newblock \showarticletitle{Domain randomization for transferring deep neural networks from simulation to the real world}. In \bibinfo{booktitle}{\emph{2017 IEEE/RSJ international conference on intelligent robots and systems (IROS)}}. IEEE, \bibinfo{pages}{23--30}.
\newblock


\bibitem[\protect\citeauthoryear{Towers~et al.}{Towers~et al.}{2023}]%
        {towers_gymnasium_2023}
\bibfield{author}{\bibinfo{person}{Mark Towers~et al.}} \bibinfo{year}{2023}\natexlab{}.
\newblock \bibinfo{title}{Gymnasium}.
\newblock
\newblock
\urldef\tempurl%
\url{https://doi.org/10.5281/zenodo.8127026}
\showDOI{\tempurl}


\bibitem[\protect\citeauthoryear{Wei and Ying}{Wei and Ying}{2021}]%
        {wei2020fork}
\bibfield{author}{\bibinfo{person}{Honghao Wei} {and} \bibinfo{person}{Lei Ying}.} \bibinfo{year}{2021}\natexlab{}.
\newblock \showarticletitle{FORK: A Forward-Looking Actor For Model-Free Reinforcement Learning}.
\newblock  (\bibinfo{year}{2021}).
\newblock


\bibitem[\protect\citeauthoryear{Zhang, Torabi, Guan, Ballard, and Stone}{Zhang et~al\mbox{.}}{2019}]%
        {zhang19leveraging}
\bibfield{author}{\bibinfo{person}{Ruohan Zhang}, \bibinfo{person}{Faraz Torabi}, \bibinfo{person}{Lin Guan}, \bibinfo{person}{Dana~H. Ballard}, {and} \bibinfo{person}{Peter Stone}.} \bibinfo{year}{2019}\natexlab{}.
\newblock \showarticletitle{Leveraging Human Guidance for Deep Reinforcement Learning Tasks}. In \bibinfo{booktitle}{\emph{Proceedings of the Twenty-Eighth International Joint Conference on Artificial Intelligence, {IJCAI-19}}}. \bibinfo{publisher}{International Joint Conferences on Artificial Intelligence Organization}, \bibinfo{pages}{6339--6346}.
\newblock
\urldef\tempurl%
\url{https://doi.org/10.24963/ijcai.2019/884}
\showDOI{\tempurl}


\end{thebibliography}

\newpage
\appendix
\section{Requirement Specifications}
We provide the list of competency requirement specifications used to construct the CPTs for each SEBN used in the paper. 
\\
\\
\begin{small}

\noindent
DoorKey
\begin{verbatim}

goalreached : (distance=1| move=2) 
goalreached : (wall=1 | move=1, avoid_wall=1)
goalreached : (exists_door=1 | haskey=1, move=1)
goalreached : (wall=1,  exists_door=1 | dooropened=1, 
                                        avoid_wall=1, move=1)
dooropened : (exists_door=1 | haskey=1, move=1, open door=1)
haskey : (exists_door=1 |pick_up=1, move=1)
haskey : (distance=1 | move=2)

\end{verbatim}
\noindent
BipedalWalker

\begin{verbatim}

goalreached : (ground_roughness=0 | move=1)
goalreached : (ground_roughness=1 | move=1, balance=1)
goalreached : (pit_gap=1 | jump=1, balance=1, move=1)
goalreached : (pit_gap=2 | jump=2)
goalreached : (stump_height=1 | move=1, jump=1, descend=1)
goalreached : (stump_height=2 | jump=2, descend=2)
goalreached : (stair_width=1 | climb=1, descend=1)
goalreached : (ground_roughness=4 | move=2, balance=2)
goalreached : (stair_steps=2 | climb=1, descend=1)

\end{verbatim}

\noindent
Robosuite
\begin{verbatim}

goalreached : (mass=0 | move_arm=1, open_door=1)
goalreached : (lock=1 | unlock=1)
goalreached : (mass=3 | open_door=2)

\end{verbatim}
\end{small}

\newpage

\section{Approximate Inference Algorithm}
For choosing candidate environments in SEBNs with a large environment design space such as the BipedalWalker domain, we adapt the KL-search algorithm from \cite{hsiao2024surrogate}.

\begin{algorithm}
\SetAlgoLined
$T\gets$ the OR-search tree on $A$ using ordering $o$\;
$OPEN\gets \{<root(T), 0>\}$ \tcp{frontier nodes are ordered by the second value}
\For{$i = 1 \rightarrow L$}{
	$v\gets OPEN$.dequeue() \tcp{remove the node in OPEN with highest priority value}
	\For{$u \in children(v)$}{
            $P_{t}(u) \gets P(u)$ on $(X,D,\Phi)_t$\\
            $P_{t+1}(u) \gets P(u)$ on $(X,D,\Phi)_t$ updated with $\Phi_B^*$\\
		$h_{kl}(u) \gets |\log (P_{t}(u) - \log (P_{t+1}(u))| \cdot P_{t+1}(u)$\\
		Append $<u, h_{kl}(u)>$ to $OPEN$
	}
}
Let $C$ be an empty list
\For(\tcp*[f]{nodes in $OPEN$ are leaves}){$v \in OPEN$}{
	Forward sample $e$, a full configuration of all environment variables in $E$ conditioned on the partial configuration represented by $v$\\
	Append $e$ to $C$
}
Return $C$
 \caption{Candidate Selection for SEBN-guided Automated Curriculum\newline
 \textbf{Input:} An SEBN $(X,D,\Phi)_t$, solution to (Equation \ref{eq:mle}) $\Phi_B^*$, a variable ordering $o$ over environment variables $E \in X$ \newline
 \textbf{Parameters:} Number of samples $L$ \newline
 \textbf{Output:} $L$ candidate environments }
 \label{alg:candidate_select}
\end{algorithm}

\section{DoorKey Details}
\begin{figure}
    \includegraphics[width=0.45\textwidth]{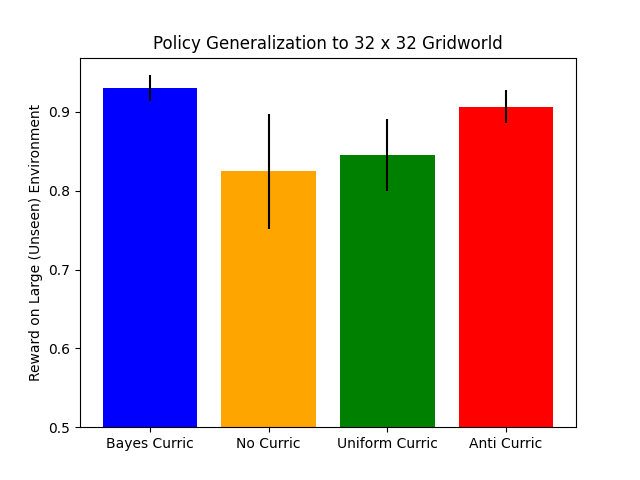}
    \caption{Reward for learned policies from Fig. \ref{fig:robosuiteres} on an unseen large gridworld environment (size 32 x 32). 
    }
    \label{fig:doorkeylarge}
\end{figure}

We provide an additional plot for policy transfer onto a large (32 x 32) gridworld environment in Fig. \ref{fig:doorkeylarge}. 

\section{Bipedal Walker Details}
\begin{figure}
    \includegraphics[width=0.40\textwidth]{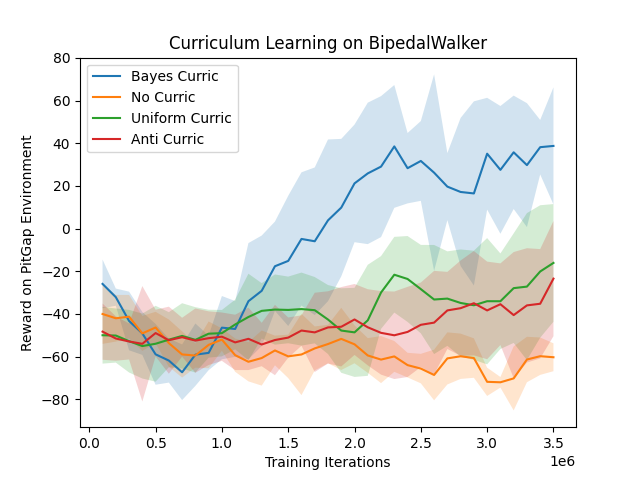}
    \includegraphics[width=0.40\textwidth]{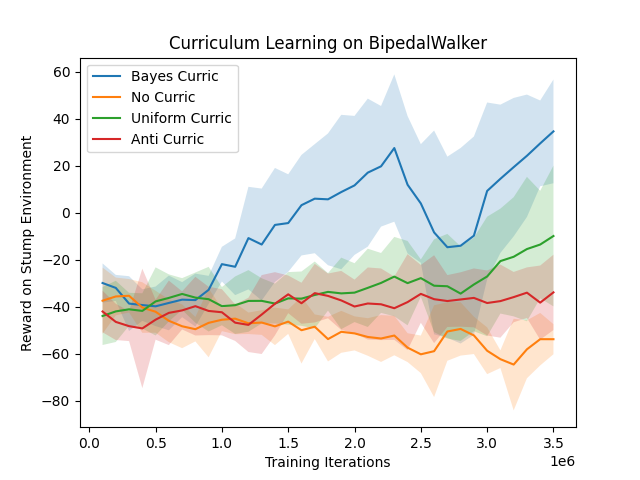}
    \includegraphics[width=0.40\textwidth]{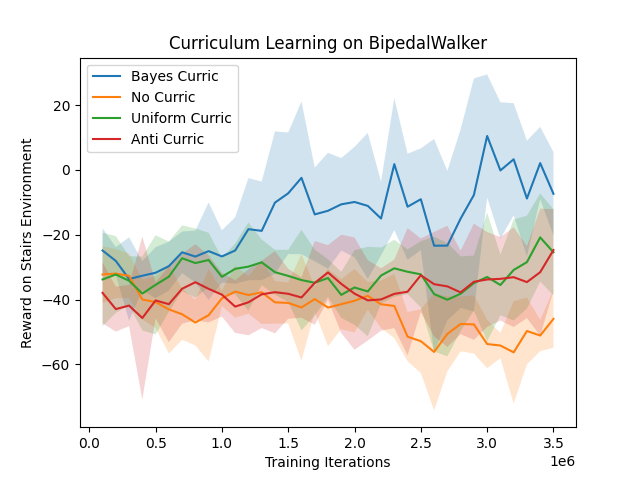}
    \includegraphics[width=0.40\textwidth]{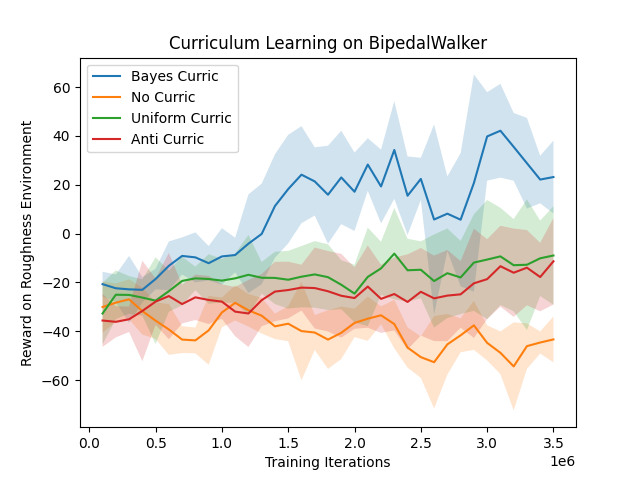}
    \caption{SEBN-guided automated curriculum on the BipedalWalker over 4 individual challenges (the four smaller plots). Evaluation environments are randomly generated within a given environment feature set.
    }
    \label{fig:bipedalwalkermulti}
\end{figure}

We provide plots for each of the individual challenges (PitGap, Stump, Stairs, and Roughness) in Fig. \ref{fig:bipedalwalkermulti}. 

\section{Robosuite Details}
\begin{figure}[b]
    \includegraphics[width=0.4\textwidth]{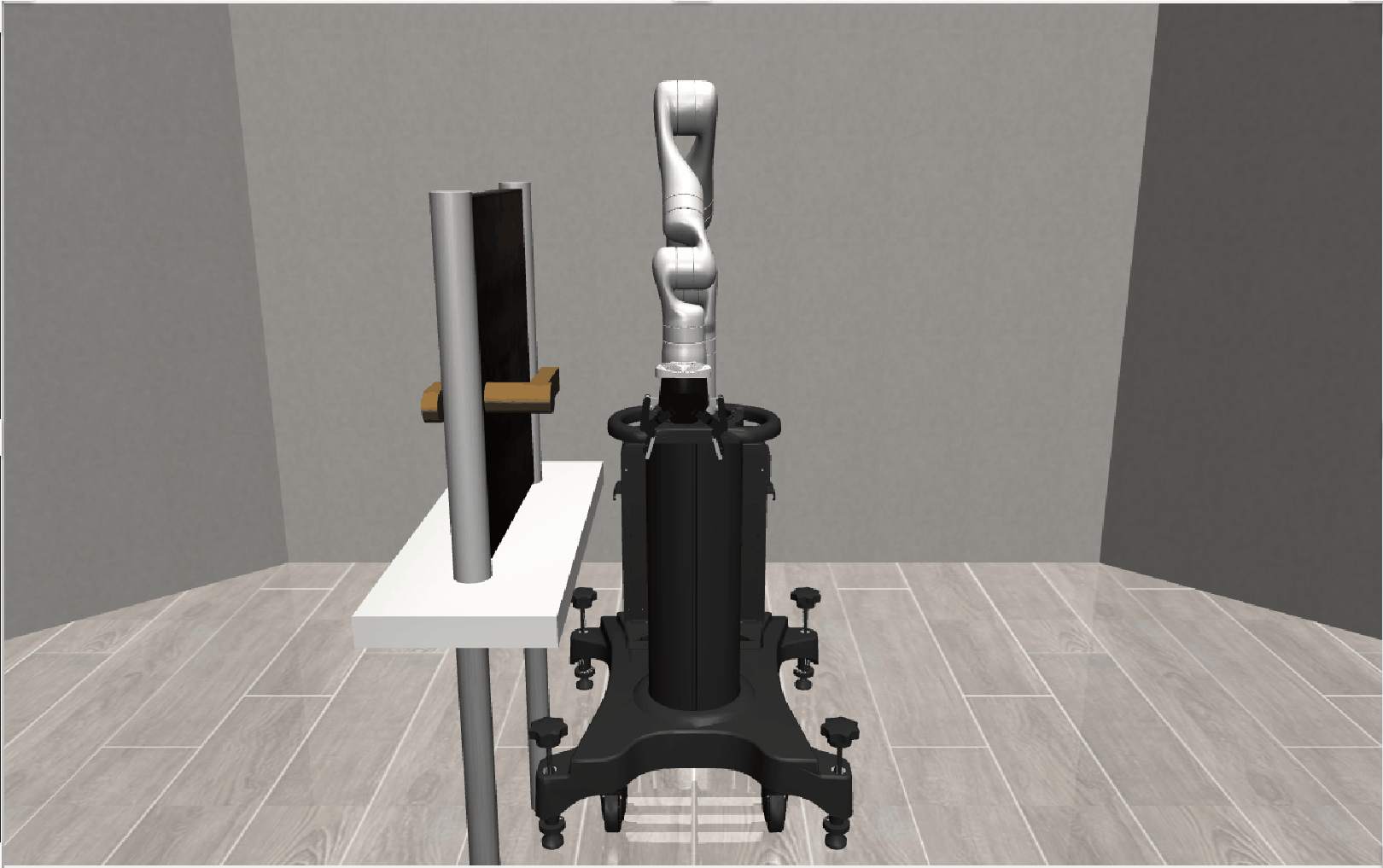}
    \caption{Door task from Robosuite simulation environment. }
    \label{fig:robosuiteenv}
\end{figure}

An example of the task environment for the Robosuite Door environment can be seen in Fig. \ref{fig:robosuiteenv}. In the example image, the latch setting is turned on, requiring the robot to apply force to rotate the handle before being able to open the door. We also provide an additional graph for transferring policies to a heavy door environment with a mass setting of 12 (where the maximum setting in the curriculum was 6). 

\begin{figure}
    \includegraphics[width=0.45\textwidth]{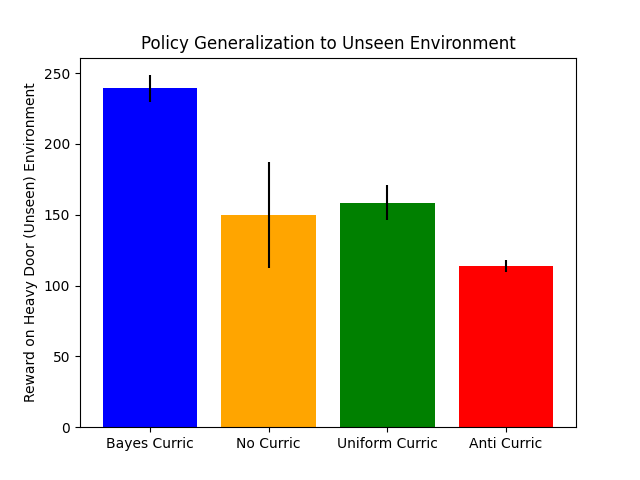}
    \caption{Reward for learned policies from Fig. \ref{fig:robosuiteres} on an unseen heavy door environment with a mass setting of 12.
    }
    \label{fig:robosuiteheavy}
\end{figure}

\end{document}